\documentclass[sigconf]{acmart}

\usepackage{booktabs}
\usepackage{cleveref}
\usepackage{dsfont}
\usepackage{gensymb}
\usepackage{hyperref}
\usepackage{listings}
\usepackage[listings, skins, most]{tcolorbox}
\usepackage{makecell}
\usepackage{multirow}
\usepackage{subcaption}
\usepackage[dvipsnames]{xcolor}
\usepackage{xspace}

\AtBeginDocument{%
  }

\copyrightyear{2026}
\acmYear{2026}
\setcopyright{cc}
\setcctype{by}
\acmConference[CAIS '26]{ACM Conference on AI and Agentic Systems}{May 26--29, 2026}{San Jose, CA, USA}
\acmBooktitle{ACM Conference on AI and Agentic Systems (CAIS '26), May 26--29, 2026, San Jose, CA, USA}
\acmDOI{10.1145/3786335.3813163}
\acmISBN{979-8-4007-2415-2/2026/05}

\newcommand{\system}{\textsc{AgentStop}\xspace}

\begin{document}

\title{\system: Terminating Local AI Agents Early to Save Energy in Consumer Devices}

\author{Dzung Pham}
\authornote{Work done during internship at Brave Software.}
\affiliation{%
  \institution{University of Massachusetts Amherst}
  \city{Amherst}
  \state{Massachusetts}
  \country{USA}
}
\email{dzungpham@cs.umass.edu}

\author{Kleomenis Katevas}
\affiliation{%
  \institution{Brave Software}
  \city{London}
  \country{UK}}
\email{kkatevas@brave.com}

\author{Ali Shahin Shamsabadi}
\affiliation{%
  \institution{Brave Software}
  \city{London}
  \country{UK}
}
\email{ashahinshamsabadi@brave.com}

\author{Hamed Haddadi}
\affiliation{%
  \institution{Brave Software, Imperial College London}
  \city{London}
  \country{UK}
}
\email{h.haddadi@imperial.ac.uk}

\renewcommand{\shortauthors}{Pham et al.}

\begin{abstract}
Autonomous agents powered by large language models (LLMs) are increasingly used to automate complex, multi-step tasks such as coding or web-based question answering. While remote, cloud-based agents offer scalability and ease of deployment, they raise privacy concerns, depend on network connectivity, and incur recurring API costs. Deploying agents locally on user devices mitigates these issues by preserving data privacy and eliminating usage-based fees. However, agentic workflows are far more resource-intensive than typical LLM interactions. Iterative reasoning, tool use, and failure retries substantially increase token consumption, often expending significant compute without successfully completing tasks.

In this work, we investigate the time, token, and energy overhead of locally deployed LLM-based agents on consumer hardware. Our measurements show that agentic execution increases GPU power draw, temperature, and battery drain compared to single-inference workloads. To address this inefficiency, we introduce \system, a lightweight efficiency supervisor that predicts and preemptively terminates trajectories unlikely to succeed. Leveraging low-cost execution signals, such as token-level log probabilities, \system can reduce wasted energy by 15-20\% with minimal impact on task performance (<5\% utility drop) for challenging web-based question answering and coding benchmarks. These findings position predictive early termination as a practical mechanism for enabling sustainable, privacy-preserving LLM agents on user devices.
Our project code and data are available at \url{https://github.com/brave-experiments/AgentStop}.

\end{abstract}



\maketitle

\section{Introduction}

Autonomous agents powered by large language models (LLMs) hold great promise for automating a wide range of user tasks, including writing and debugging code, conducting online shopping, making bookings, and answering questions.
In a common deployment paradigm, such agents are offered as cloud-hosted services, where a provider exposes a natural language interface through which users issue instructions and supply the data to be processed, while the underlying LLM reasoning, acting, and environment interaction is executed remotely on the provider's infrastructure. Yet, this architecture introduces three fundamental challenges: privacy risks, practical constraints, and financial costs.

Consider, for example, a coding task such as automated bug fixing. The source code submitted by the user (the primary data to the agent) often contains sensitive information, including file paths, directory structures, and proprietary business logic, making it valuable intellectual property. Transmitting such artifacts to a third-party agent provider risks exposing confidential data, potentially violating privacy obligations, and compromising competitive advantage. Beyond privacy, this submission requires network accessibility, and given the size and complexity of real-world codebases, repeated transmissions across extended agentic workflows introduce significant bandwidth consumption and cumulative latency. Finally, agentic tasks are inherently financially demanding: unlike typical LLM interactions that involve single prompt-response exchange, autonomous agents repeatedly run model inferences and call external tools across multi-step reasoning chains. Executing the above coding task alone costs $\approx$1~USD~\citep{jimenez2024swebench}, and at scale, sustained deployments can translate into tens of thousands of dollars in monthly operating expenses~\cite{openai_api_pricing_2026, anthropic_claude_pricing_2026}.

Local agents address these challenges by operating locally on user devices: they protect privacy by not sharing any data with the agent provider, eliminate dependence on external infrastructure, and reduce financial costs. However, relocating computation to user devices introduces a new constraint: sustained resource consumption on user devices. In these cases, long agentic workloads can lead to significant battery drain, thermal pressure, and device wear. Unlike server-side computation, where energy costs are outsourced to data centers, on-device execution directly competes with users’ everyday needs, potentially degrading device availability and usability. In practice, agents amplify model usage through long action sequences, frequent decision points, and tool interactions, leading to resource demands that are substantially higher than those suggested by per-inference measurements alone. As a result, the impact of LLM-based agents on system resources, particularly battery life, can be far more severe than anticipated.

Resource pressure from on-device agents also has usability implications. Prior work has documented \emph{nomophobia} (i.e., ``no-mobile-phone phobia''), an anxiety associated with losing access to a functioning mobile device, which is often exacerbated by low battery levels~\cite{yildirim2015exploring}. Such concerns correlate with usage patterns such as persistent background activity and reluctance to power off devices~\cite{katevas2018typical}. An agent that rapidly depletes battery life increases anxiety and discourages sustained adoption in mobile or always-on scenarios.

\begin{figure*}
    \centering
    \includegraphics[width=\linewidth]{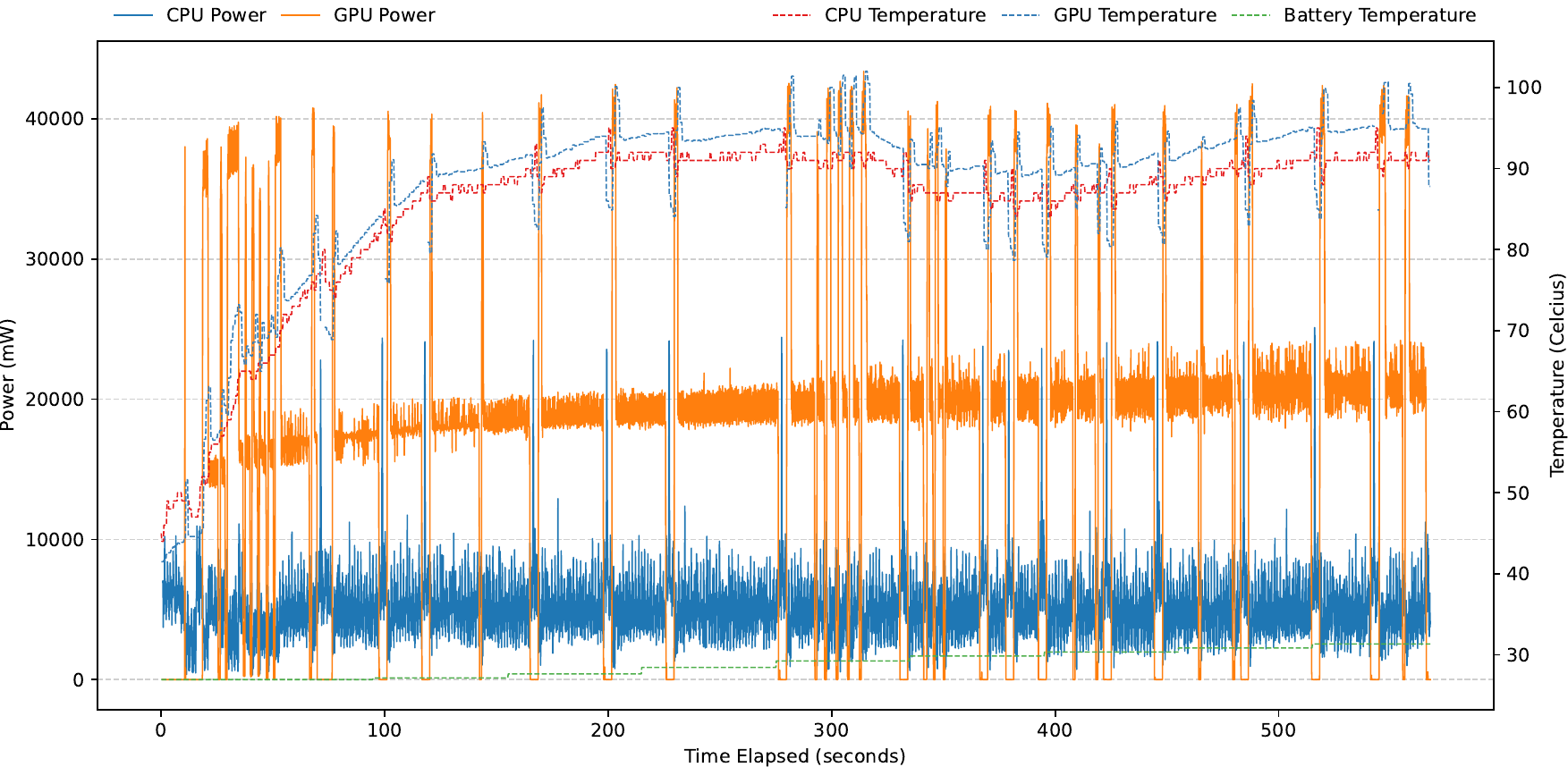}
    \caption{Profile of instantaneous power (left y-axis) and hardware temperature (right y-axis) over time (x-axis) on an Apple M1 Max laptop for a coding agent powered by Qwen3-Coder-30B-A3B~\citep{alibaba2025qwen3} solving a SWE-Bench Verified task~\citep{jimenez2024swebench}. Instantaneous GPU power gradually increases alongside CPU/GPU/battery temperature. GPU power spikes correspond to the pre-filling stage of LLM inference, while the lower but more sustained patterns correspond to the generation stage.}
    \label{fig:power_and_temp}
\end{figure*}

In this work, we investigate the resource implications of locally deployed LLM-based agents and introduce a practical mechanism to mitigate their inefficiencies. We empirically characterize the time and energy overhead incurred by agentic execution on a consumer-grade device (e.g., MacBook Pro M1 Max). In contrast to a typical LLM inference, we show that the iterative execution, including failure retry loops, substantially amplifies token consumption, latency, and battery drain, often without producing a successful task outcome.
Figure~\ref{fig:power_and_temp} shows an example agentic coding task lasting $\approx$600s with 30+ LLM inference calls and terminal tool calls, during which GPU power repeatedly spikes to more than 40W. GPU temperature also rises to nearly 95\degree C, with sustained operation above 90\degree C, indicating extended thermal stress.

To address this inefficiency, we propose \system, a lightweight efficiency supervisor designed to predict and preemptively terminate agent runs that are unlikely to succeed. The supervisor can be deployed at different levels: (a) users may implement a local early-stopping mechanism to control cost and resource usage, or (b) agent providers may offer it as a built-in feature to proactively prevent unnecessary computation and reduce user expenses. We formulate early termination as a binary supervisor prediction problem and train a gradient-boosted decision tree-based model on lightweight features directly extracted from agent execution traces.
These include token-level features such as log probabilities, generated during reasoning and tool use.
Importantly, \system relies exclusively on signals that are already produced during standard inference, introducing negligible additional computation or energy overhead, and requires no modification to the underlying model.

We evaluate \system on representative agent workloads, including web-based question answering and terminal-based coding benchmarks. Our results demonstrate that \system can reduce energy wastage of a Qwen3-30B-A3B-powered agent by 15--20\% with <5\% task utility drop on challenging question answering and coding benchmarks.
Beyond the performance improvements, our work reframes efficiency supervision as a key enabler of sustainable, privacy-preserving, on-device agents. By mitigating unnecessary battery drain and execution time, predictive early termination not only improves system-level resource usage but also addresses important human-centered concerns surrounding device longevity, usability, and trust.

To summarize, our contributions include:

\begin{itemize}
    
    \item We introduce \system, a lightweight ML-based supervisor that predicts unsuccessful agent trajectories using signals already available during agent execution.

    \item We demonstrate that predictive early termination reduces energy consumption and execution time across representative web and coding agent benchmarks.
    
    \item Beyond system-level gains, we position efficiency supervision as a key enabler for privacy-preserving, on-device agents by mitigating battery drain, improving usability, and reducing barriers to real-world adoption.
    
\end{itemize}

\section{Background and Related Work}

In this section, we provide an overview of local LLM-powered agents and their energy usage as well as energy-saving strategies.

\subsection{Local Agents}

AI agents are autonomous software systems that leverage AI technology, particularly large language models (LLMs), to solve tasks for users.
An agent execution often alternates between LLM inference---a GPU-intensive phase where the reasoning and planning occur---and environment interaction, which involves calling specialized external tools (e.g., web search).
State-of-the-art agents are usually powered by cloud-based LLMs (e.g., Claude Code\footnote{\url{https://code.claude.com/docs/en/overview}}) that not only require a direct financial cost but also raise privacy risks.

In contrast, local AI agents use locally deployed language models that ``fit onto a common consumer electronic device''~\citep{belcak2025slm}, such as mobile phones, tablets, and PC/laptops with 2--16 GB of memory (RAM) or up to 32--64 GB in higher-end machines (e.g., MacBook Pro/Max series).
Any user requests to these local agents stay completely on-device with minimal delay.
This setup thus enables better privacy protection and accessibility, but at the expense of task performance and high energy usage due to hardware limitations~\citep{lu2025demystify}.

To improve the efficiency of running local language models, a popular technique is to apply quantization to the models' weights (and key-value cache), yielding a reduction of up to 2--4x in memory footprint, with a small impact on utility~\citep{laskaridis2024melt, li2025palmbench}.
Another increasingly common technique is the Mixture-of-Experts (MoE) model architecture, where instead of always using all parameters, the model is trained to select a small subset of the parameters to perform the inference~\citep{shazeer2017moe}.
By combining quantization with MoE, we can more efficiently run LLMs with as many as 30 billion parameters completely on consumer devices equipped with 24GB RAM.

\subsection{LLM Energy Usage in Consumer Devices}

Despite the use of these efficiency optimizations, LLMs still impose a non-negligible amount of stress on consumer devices.
For example, measurement frameworks like MELT~\citep{laskaridis2024melt} or PalmBench~\citep{li2025palmbench} find that running a Gemma 2B model on an iPhone 14 Pro can take $\approx$3 mAh per 100 output tokens, which is roughly equivalent to using $\approx$ 0.1\% of the device's battery to generate about 60-80 English words or 4-5 English sentences.
Even though this cost may seem small at first glance, agentic workloads like coding often require (tens of) thousands of tokens to be generated, not to mention the input tokens to be processed, which in real-world information-seeking scenarios is often an order of magnitude more than the output tokens.
And while local LLM deployment has seen impressive efficiency gains on non-agentic tasks (2--3x intelligence per watt year-over-year~\citep{saadfalcon2025ipw}), the efficiency gap between local and cloud deployment remains large~\citep{google2024energy, google2025measuring, saadfalcon2025ipw}.
Moreover, most agent benchmarks do not directly measure energy consumption, often only using proxy metrics such as the number of tokens and latency~\citep{zhu2025establishing}.
\citep{kim2025cost} is the only work that investigates agents from the viewpoint of efficiency, but only in the cloud server environment and without considering energy usage.

\subsection{Early Stopping Strategies for Efficiency}

In natural language processing, one simple yet effective approach to improve model inference efficiency is to stop the process early~\citep{zhou2020bertexit, sun2022hashexit} based on some form of confidence or uncertainty signals~\citep{lin2024uncertainty}.
The idea has also found applications in LLM cascades, where inference starts with a weaker LLM and may be cascaded to a stronger one or stopped~\citep{yue2024llmcascade, gupta2024lmcascade, zellinger2025earlyabstention}.
To the best of our knowledge, early exit has only been studied in simple, single-turn LLM inference and has not been applied to the multi-turn agent settings.
\citep{lu2025runaway} is the closest work that mainly studies in ``intrinsic early exit'', where the agents can select an EXIT action to stop themselves.
The paper only evaluates on well-structured benchmarks with trackable progress and does not measure energy savings.

\section{Problem Statement}

Consider an agent $\mathcal{A}$ and its execution history $\mathcal{H}$ consisting of sequences of environment states $S_i$ and agent actions $A_i$, e.g., $\mathcal{H}\allowbreak =\allowbreak [S_1,\allowbreak A_1,\allowbreak S_2,\allowbreak A_2,\allowbreak \ldots,\allowbreak S_t,\allowbreak A_t]$.
We define our efficiency supervisor as a classifier $\mathcal{C}$ that takes $\mathcal{H}$ as inputs and returns the probability that $\mathcal{A}$ can proceed from $\mathcal{H}$ and successfully execute its task, e.g. $C: \mathcal{H} \rightarrow [0, 1]$.
In the case of LLM agents, $S_1$ consists of the system and initial task prompt for the agent $\mathcal{A}$, each subsequent $S_i$ consists of the outputs from tool calls made by $\mathcal{A}$, and each action $A_i$ consists of the generated tokens along with their log-probabilities (logprobs).
Thus, to create the supervisor $\mathcal{C}$, we can adopt supervised learning by collecting multiple different trajectories  $\mathcal{H}_i$ along with their final outcome $O_i \in \{0, 1\}$ (0 means failure, 1 means success), then train $\mathcal{C}$ on (parts of) these trajectories to predict the outcomes.

Given a list of incomplete $\mathcal{H}_i$ with the final outcome $O_i$, to measure the effectiveness of early stopping with $\mathcal{C}$ and a prediction threshold $\tau$, we propose the following metrics:
\begin{itemize}
    \item Energy wastage: Energy spent by the agent on failed runs:
    $\sum_i (\text{Energy}(\mathcal{H}_i) \cdot \mathds{1}{\{O_i = 0\}})$
    \item Early-stop energy wastage:
        \[
\sum_i 
\Bigg(
\begin{aligned}
&\phantom{....} \text{Energy}(\mathcal{H}_{i}^{+}) \cdot 
\mathds{1}{\{O_i = 0 \wedge C(\mathcal{H}_i) \ge \tau\}}
\\
&+ \text{Energy}(\mathcal{H}_i) \cdot 
\mathds{1}{\{O_i = 0 \wedge C(\mathcal{H}_i) < \tau\}}
\\
&+ \text{Energy}(\mathcal{H}_i) \cdot 
\mathds{1}{\{O_i = 1 \wedge C(\mathcal{H}_i) < \tau\}}
\\
&+ \text{Energy}(C(\mathcal{H}_i))
\end{aligned}
\Bigg)
\]
        where $\text{Energy}$ measures the energy of the input action and $\mathcal{H}_{i}^{+}$ represents the completed trajectory.
        Essentially, this metric sums four terms: (1) energy used by failed runs that were not stopped early, (2) energy used by failed runs that were stopped early, (3) energy used by would-be successful runs that were turned into failures due to early stopping, and (4) the energy used by the classifier $\mathcal{C}$.

    \item Energy wastage reduction percentage: Measures the percentage of energy wastage reduced by early stopping:
        \[\Big(1 - \frac{\text{Early stop energy wastage}}{\sum_{i} \text{Energy}(\mathcal{H}_{i}^{+}) \cdot \mathds{1}{\{O_i = 0\}} } \Big) \cdot 100\%\]
    \item Task utility drop: Measures the drop in task utility due to early stopping:
        \[
          \frac{\sum_{i} \mathds{1}{\{O_i = 1 \wedge \mathcal{C}(\mathcal{H}_i) < \tau\}}}{\sum_{i} O_i} \cdot 100\%
        \]
\end{itemize}

Measuring the energy consumption provides a more meaningful picture of agents' real-world impact than simply tracking token usage or latency, since they do not capture the underlying computational intensity, hardware utilization, cooling overhead, or carbon footprint of running models at scale.
Two systems may generate the same number of tokens with similar latency while consuming vastly different amounts of electricity due to architectural differences, batching strategies, or hardware choices.
Furthermore, as can be seen in Figure \ref{fig:power_and_temp}, an agent's energy usage can increase due to sustained use.
Using energy-based metrics also means we need to build the supervisor $\mathcal{C}$ in a way that its energy cost does not outweigh its gains.
For example, implementing $\mathcal{C}$ using large, GPU-reliant machine learning models like an LLM could potentially offset any benefits from the increased capabilities.
Considering the constraints of local deployment on consumer hardware, this requirement for a lightweight supervisor is thus critical.

\section{Building \system}

To create the efficiency supervisor, we adopt a data-driven approach that leverages easily extractable features from the agent execution traces to train a classifier that can efficiently predict whether the agent will succeed or not.

\subsection{Collecting Training Data}

We start with collecting a small, labeled dataset of both successful and failed attempts by the agent at solving tasks, particularly web-based question answering and coding.
Each data record contains the following information:
\begin{itemize}
    \item All generated tokens and their log probabilities (logprobs): Each token generated by the agent's LLM is associated with a logprob value that can be retrieved from the underlying inference engine. This logprob represents the likelihood of the token being chosen and is also indicative of the LLM's ``confidence''~\citep{yue2024llmcascade, gupta2024lmcascade}.
    \item All tool calls made by the agent and their outputs: The agent always makes at least one tool call in every iteration of their ReAct loop. The employed tools may encounter errors.
    \item Success/Failure label: Whether or not the agent successfully completed the task.
\end{itemize}

This collection of data can be readily created by evaluating the agent on existing benchmarks such as SWE-Bench~\citep{jimenez2024swebench}, using popular local LLM inference frameworks that support logprobs retrieval (e.g., LlamaCpp~\citep{llamacpp}).
Depending on the use case, the dataset can be collected just once or can be continually updated with new records labeled by the user.

\subsection{Feature Engineering}

We extract the following features from each agent trajectory:
\begin{itemize}
    \item Top $k$ smallest output logprobs at each agent step: Unlike prior work that only looks at a single smallest logprobs or at a specific quantile~\citep{gupta2024lmcascade}, we are interested in the ``tail'' of the agent's confidence throughout all of its interactions. We further exponentiate these logprobs to keep their range in $[0, 1]$. $k$ can be a tunable hyperparameter, but for our experiment, we choose $k=10$ for simplicity.
    \item Number of output tokens at each step: The length of the agent's chain-of-thought (CoT) and action (e.g., a code patch) may correlate with the difficulty of the task. For example, a long CoT may indicate that the agent is struggling.
    \item Ratio of tokens overlap between any two adjacent agent steps: LLM-based agents can sometimes repetitively attempt the same actions, which may indicate that the agents have fallen into a loop. We measure this overlap by calculating the length of the longest common sequence between the two steps' tokens and dividing this by the number of tokens in the preceding step.
\end{itemize}

\subsection{Modeling and Deployment}

Once the data is ready, we train a gradient-boosted decision trees (GBDT) model using XGBoost~\citep{chen2016xgboost} with stratified nested 5-fold cross-validation.
The outer folds are used to obtain more robust test results, while the inner folds are for performing hyperparameter optimizations.
Using the halving randomized search strategy (more trees as search continues), we tune various hyperparameters such as learning rate and max tree depths (\Cref{apd:hyperparams}).

There are three main reasons why we favor GBDT over other machine learning algorithms like simple linear regressions or neural nets: a) GBDT's inference latency and energy consumption are low, using less than $0.01$ mWh per inference~\citep{santos2024green}; b) GBDT generally exhibits strong performance on structured, tabular data~\citep{grinsztajn2022tab, mcelfresh2023tab}; and c) training and tuning GBDT are also efficient.
For the experiments, we train one GBDT per combination of agent, target task, and agent step.
For a given agent run, the selected GBDT model will only run once after the associated agent step is finished to determine if the agent should continue or not.

\section{Experimental Setup}

Here, we describe our evaluation settings, particularly the chosen benchmarks, agent frameworks, metrics, and baselines.

\subsection{Benchmark Tasks}

\subsubsection{Question answering (QA)}

We test our agents on two popular QA datasets: SimpleQA~\cite{wei2024simpleqa} and FRAMES~\citep{krishna2025frames}.
SimpleQA consists of 4,326 ``simple'' questions that can theoretically be answered using exactly one reference document, whereas FRAMES consists of 824 questions that require multi-hop reasoning over multiple documents.
We provide our agents with internet access to find these references, including web search with the Brave Search API.\footnote{\url{https://brave.com/search/api/}}
We exclude any questions to which the underlying LLMs already know the answers without internet access, since we only want to target unseen questions.
To evaluate the correctness of the agents' answers, we prompt the Claude Haiku 4.5 model with OpenAI's evaluation prompt for SimpleQA.

\subsubsection{Coding}
We use the SWE-Bench Verified dataset~\citep{jimenez2024swebench}, which consists of 500 GitHub issues that have been human-verified for feasibility.\footnote{\url{https://openai.com/index/introducing-swe-bench-verified/}}
The agents are given access to the relevant GitHub repositories via Docker containers and are tasked with resolving the mentioned GitHub issues.
We allow the agents to run arbitrary Bash commands inside the Docker containers and disable all internet access to be consistent with the official Bash-only leaderboard by the developers of SWE-Bench.
The agent outputs in this case are code patches generated by ``git diff'' and evaluated against a set of tests inside Docker containers.

\subsection{Agent Implementation}

\subsubsection{Architecture and Harness}
We adopt the CodeAct agent architecture~\citep{wang2024codeact}, which essentially follows the ReAct loop~\citep{yao2023react} but generates code for the actions instead.
For the web-based QA task, we use the default agent harness from HuggingFace's smolagent framework~\citep{smolagents} with slightly modified tools to format web pages to be cleaner.
smolagent comes with built-in instrumentation code, which allows us to obtain detailed information about each step and action taken by the agents.
For each question, we allow the agents to run for at most 10 steps with no more than 512 output tokens per step and a 40960 max context window length.
If the agents fail to emit a final result by the end of the 10 steps, they are forced to do so in an 11-th step using all the contexts so far.

For coding, we reuse smolagent's harness but replace all system prompts and task instructions with those from the mini-swe-agent (v1) framework~\citep{yang2024sweagent}, which has been customized for Bash-only evaluation on SWE-Bench.
Due to the difficulty of this task, we extend the maximum number of steps to 100 with at most 4096 output tokens per step and a 81920 max context window length.
Similar to the QA, a 101-th step may be used to force the agent to produce a final answer if the first 100 steps fail to do so.

\subsubsection{LLMs and Inference Backend}
We primarily use models from the Qwen 3 series~\citep{alibaba2025qwen3}, focusing on the Qwen-3-30B-A3B-2507-Instruct model for QA and the Qwen-3-Coder-30B for coding.
These two mixture-of-expert models have consistently demonstrated state-of-the-art performance for their 30-billion-parameter weight class while only using as much energy as a 3-billion-parameter model.
Importantly, they can be used in non-reasoning mode, which is crucial to efficient local deployment since reasoning models tend to generate a large volume of thinking tokens.
We use the 4-bit quantized checkpoints from Ollama, which require at least 24 GiB of video memory.
However, because Ollama does not provide access to the logprobs, we use Llama.cpp instead as the main inference backend to run these checkpoints.
We use the same decoding parameters (e.g., temperature, top-p, min-p) as recommended by the Qwen team.

\subsection{Energy Measurement}
\label{sec:energy_measurement}

We perform our evaluations on an Apple Silicon M1 Max laptop (macOS 15.6.1) equipped with 64 GiB of unified memory and a 24-core integrated GPU.
The M1 Max represents a relatively high-end personal computing device suitable for intensive workloads such as video editing and running LLMs.
To measure power usage, we use the MacOS-native \texttt{powermetrics} tool,\footnote{\url{https://firefox-source-docs.mozilla.org/performance/powermetrics.html}} which allows us to log the instantaneous power (mW) every 100 milliseconds for different components of the laptop (e.g., CPU, GPU).
We use the trapezoidal integration rule to ``integrate'' over the power to estimate the agents' energy expenditure (mWh), which is further broken down by each agent step and action (e.g., model inference vs. tool use).
To ensure fair accounting, we subtract the laptop's baseline CPU energy expense from the agent's CPU energy during its non-GPU phases.
We do not need to do the same for the GPU because our device has been modified to run in a server-like mode without any graphical user interface, so any GPU usage would only come from the agents.
(Aside from power and energy, we also use a customized version of the \texttt{glances} monitoring tool\footnote{\url{https://github.com/nicolargo/glances}} to track various other system signals such as CPU/GPU utilization, temperature, and memory usage.)

\subsection{Baselines}

We compare \system to the following baselines:
\begin{itemize}
    \item Default: We run the original CodeAct agents to completion without any intervention.
    \item Random exit: The agents are randomly stopped at certain fixed steps.
    \item Min logprob: We use the value of the smallest logprob of the latest step to determine if the agents should be stopped or not.
    \item Mean logprob: Similar to the min logprob, we use the mean of the logprobs of the latest step to determine whether to stop the agents. This is also called the Chow mean approach~\citep{gupta2024lmcascade}.
\end{itemize}

Note that we can avoid rerunning the agents for each different baseline and only need to run exactly once per task in the vanilla setting.
This is because we have the energy usage of all agent steps/actions as well as the output tokens and logprobs, which allow us to easily simulate how the energy consumption will change if we stop the agents early.

\section{Evaluation Results for Default Agents}

Here, we analyze the energy consumption when running our agents without any intervention on web-based question answering and terminal-based coding tasks.

\subsection{Web-based QA}

\begin{table}
    \small
    \centering
    \setlength{\tabcolsep}{1pt}
    
    \caption{Summary of Qwen3-30B-A3B and Qwen3-1.7B's default agent performance on FRAMES and SimpleQA.}
    \begin{tabular}{ccccccc}
        \toprule
         Model & Dataset & \makecell{\# valid \\ runs} & \makecell{Acc.} & \makecell{Avg.\\ duration(s)} & \makecell{Avg. in/out\\ tokens}  & \makecell{Avg. energy \\ wastage (mWh)} \\
         \midrule
         30B & FRAMES & 668 & 0.62 & 61.2 & 6k/0.6k & 352.4 $\pm$ 37.1\\
         30B & SimpleQA & 3525 & 0.86 & 36.4 & 3.9k/0.2k & 278.0 $\pm$ 26 \\
         1.7B & FRAMES & 746 & 0.21 & 36.7 & 4.7k/0.5k & 98.0 $\pm$ 9.4 \\
         1.7B & SimpleQA & 4231 & 0.73 & 20.4 & 3.4k/0.2k & 44.1 $\pm$ 2.1 \\
         \bottomrule
    \end{tabular}
    \label{tab:qa_perf_summary}
\end{table}
\begin{table}
    \small
    \centering
    \setlength{\tabcolsep}{1pt}
    
    \caption{Summary of Qwen3-Coder-30B-A3B's performance with the mini-swe-agent harness on SWE-Bench Verified.}
    \begin{tabular}{cccccc}
        \toprule
         Result & Count & \makecell{Avg.\\duration (s)} & \makecell{Avg. in/out\\ tokens} & \makecell{Avg. energy \\ usage (mWh)} & \makecell{Energy\\ 95\% CI}\\
         \midrule
         Success & 94 (18.8\%) & 274.7 & 15.1k/4.5k & 1467.3 & [1179.3, 1755.3] \\
         Fail & 406 (81.2\%) & 544.2 & 24.5k/8.0k & 3004.6 & [2745.3, 3263.9] \\
         \midrule
         Overall & 500 (100\%) & 493.6 & 22.7k/7.4k & 2715.6 & [2492.2, 2938.9] \\
         \bottomrule
    \end{tabular}
    \label{tab:swebench_perf_summary}
\end{table}

\begin{figure}
\centering

\begin{subfigure}[b]{0.49\columnwidth}
    \centering
    \includegraphics[width=\textwidth]{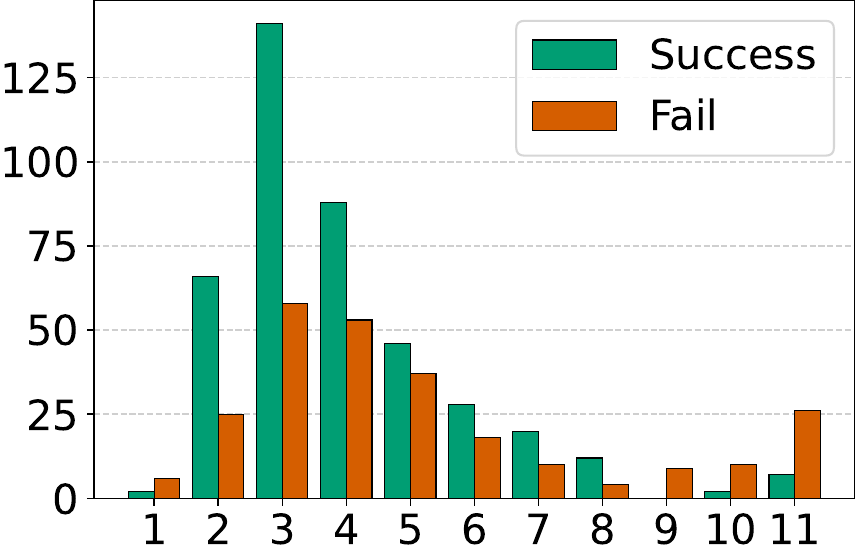}
    \caption{Qwen3-30B-A3B; FRAMES}
\end{subfigure}
\hfill
\begin{subfigure}[b]{0.49\columnwidth}
    \centering
    \includegraphics[width=\textwidth]{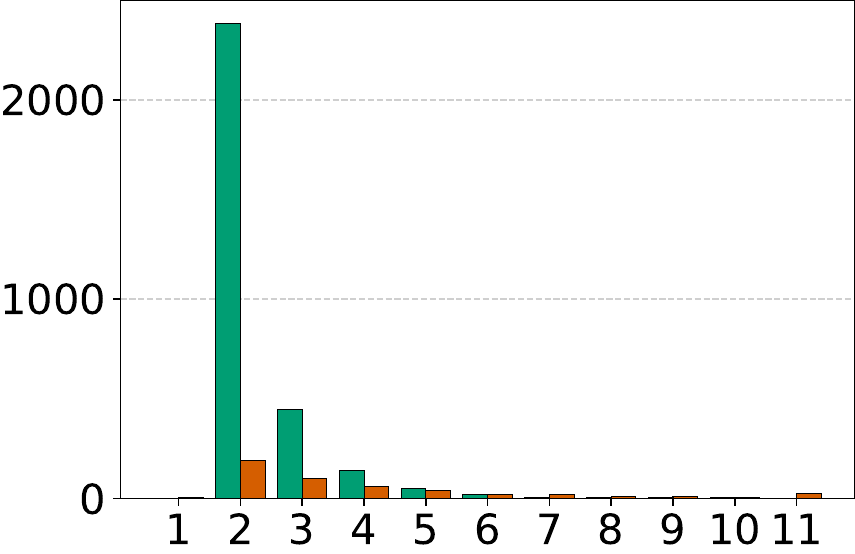}
    \caption{Qwen3-30B-A3B; SimpleQA}
\end{subfigure}

\begin{subfigure}[b]{0.49\columnwidth}
    \centering
    \includegraphics[width=\textwidth]{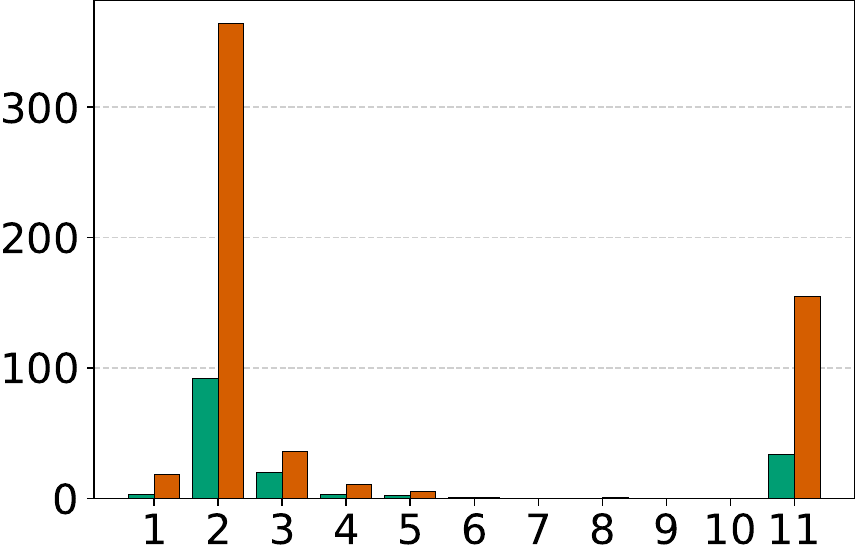}
    \caption{Qwen3-1.7B; FRAMES}
\end{subfigure}
\hfill
\begin{subfigure}[b]{0.49\columnwidth}
    \centering
    \includegraphics[width=\textwidth]{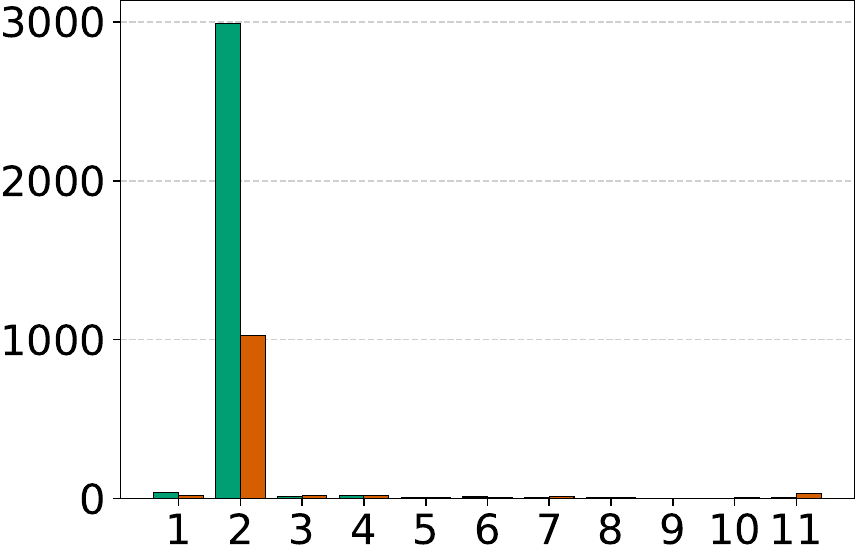}
    \caption{Qwen3-1.7B; SimpleQA}
\end{subfigure}

\centerline{\small{Number of agent steps}}

\caption{Histogram of the number of agent steps (x-axis) taken by Qwen3-30B-A3B and Qwen3-1.7B to complete each task in FRAMES and SimpleQA, split by successes/failures. For FRAMES, successful runs usually take 2--4 steps to complete, while for SimpleQA, around 2--3 steps are needed.}
\label{fig:qa_step_dist}
\end{figure}

Table \ref{tab:qa_perf_summary} presents a high-level summary of our Qwen3-powered agents on the FRAMES and SimpleQA datasets.
Compared to the 1.7B model, the 30B-A3B model achieves \textbf{nearly 3x and 1.17x accuracy} on (unmemorized) FRAMES and SimpleQA questions, respectively (while memorizing 2--9x more of the answers).
Concerningly, the MoE model \textbf{wastes 3.5--7x more energy} on average per failed task than the 1.7B version (e.g., 352.4 mWh vs 98.0 mWh per FRAMES question).
To put this in context, a modern Apple M-series laptop is often equipped with a 50--100 Wh battery, so if the MoE-powered agent fails on 10 FRAMES tasks, the total energy wastage will be roughly 3.5--7\% of the laptop's battery.
The higher success rate on SimpleQA of both models is consistent with the relatively lower difficulty of the dataset.
This is further reinforced by the distribution of the number of steps taken for each dataset (Figure \ref{fig:qa_step_dist}), where the MoE usually takes 3--4 steps to complete a FRAMES task and 2 steps for SimpleQA.

\begin{figure}
\centering
\begin{minipage}{0.03\columnwidth}
\raisebox{1cm}{
\rotatebox{90}{\small{Cumulative energy contribution (\%)}}}
\end{minipage}
\begin{minipage}{0.96\columnwidth}

\begin{subfigure}[b]{0.49\columnwidth}
    \centering
    \includegraphics[width=\textwidth]{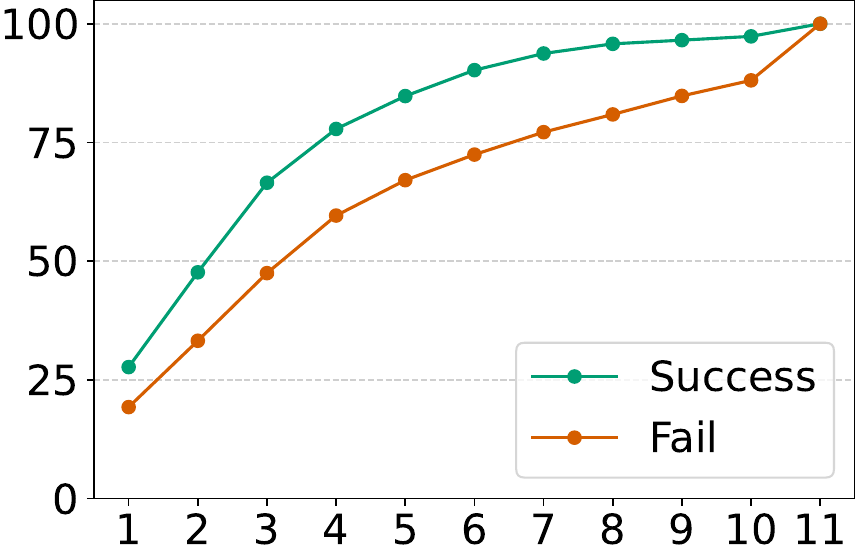}
    \caption{30B-A3B; FRAMES}
\end{subfigure}
\hfill
\begin{subfigure}[b]{0.49\columnwidth}
    \centering
    \includegraphics[width=\textwidth]{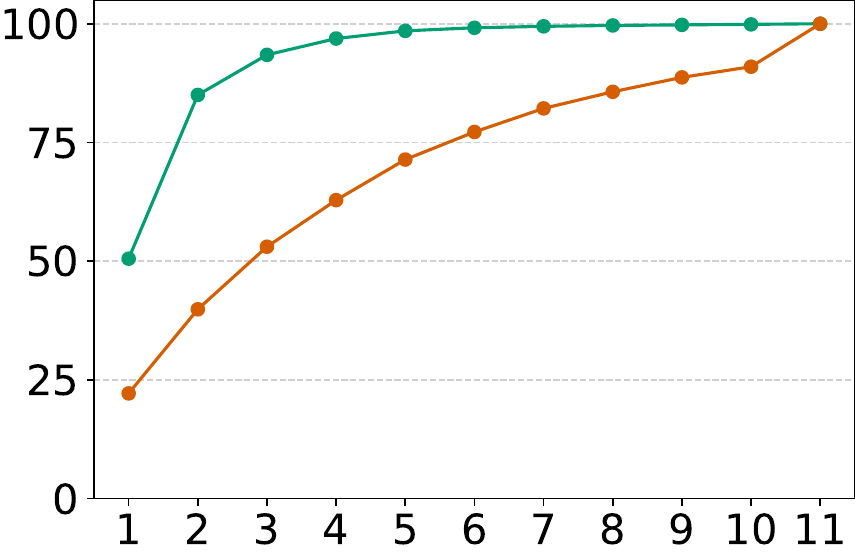}
    \caption{30B-A3B; SimpleQA}
\end{subfigure}

\vspace{2mm}

\begin{subfigure}[b]{0.49\columnwidth}
    \centering
    \includegraphics[width=\textwidth]{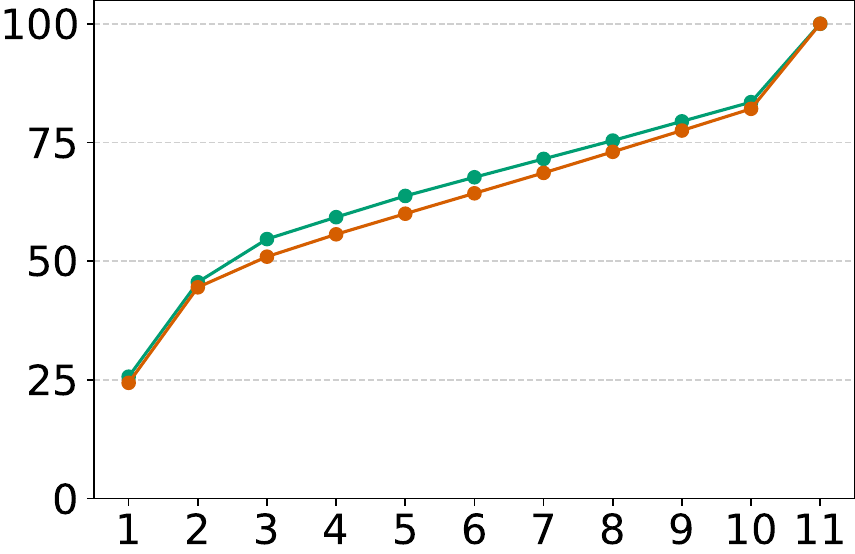}
    \caption{1.7B; FRAMES}
\end{subfigure}
\hfill
\begin{subfigure}[b]{0.49\columnwidth}
    \centering
    \includegraphics[width=\textwidth]{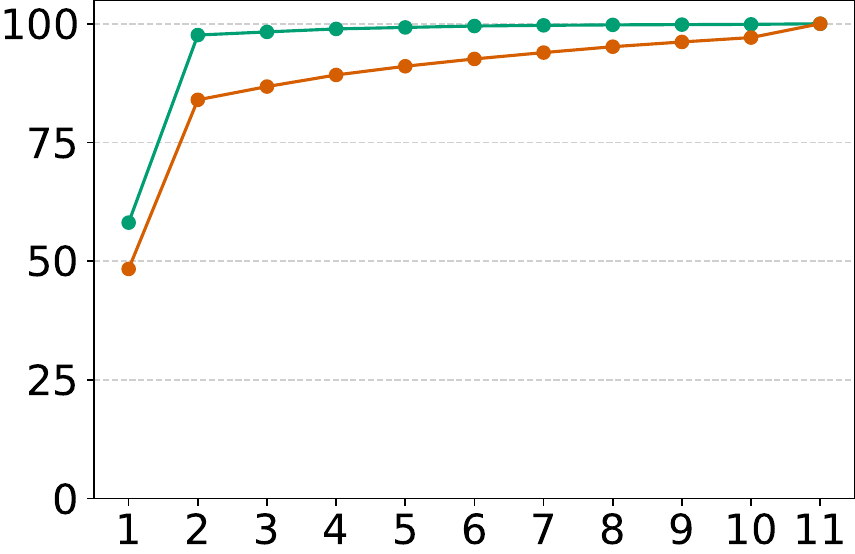}
    \caption{1.7B; SimpleQA}
\end{subfigure}

\centerline{\small{\hspace{5mm} Agent step}}

\end{minipage}

\caption{Cumulative energy contribution (\%) (y-axis) of each agent step (x-axis) for Qwen3-30B-A3B and Qwen3-1.7B on FRAMES and SimpleQA, split by successes and failures. At any fixed agent step, the energy cost of subsequent steps is higher for failed runs than for successful runs.}
\label{fig:qa_step_energy_contrib}
\end{figure}

Looking at the cumulative energy contribution of each step in Figure \ref{fig:qa_step_energy_contrib}, we observe that for any given agent step, the energy consumed up to that point by failed runs always takes a noticeably smaller proportion of their total energy consumption than does the energy consumption by successful runs.
In other words, more energy is used by the later steps in failed runs than in successful runs, which is consistent with the step count distribution in Figure \ref{fig:qa_step_dist}, where failed runs tend take longer to complete.
These observations reaffirm the importance of stopping agents early.

\begin{figure}
\centering

\begin{subfigure}[b]{\columnwidth}
    \centering
    \includegraphics[width=\textwidth]{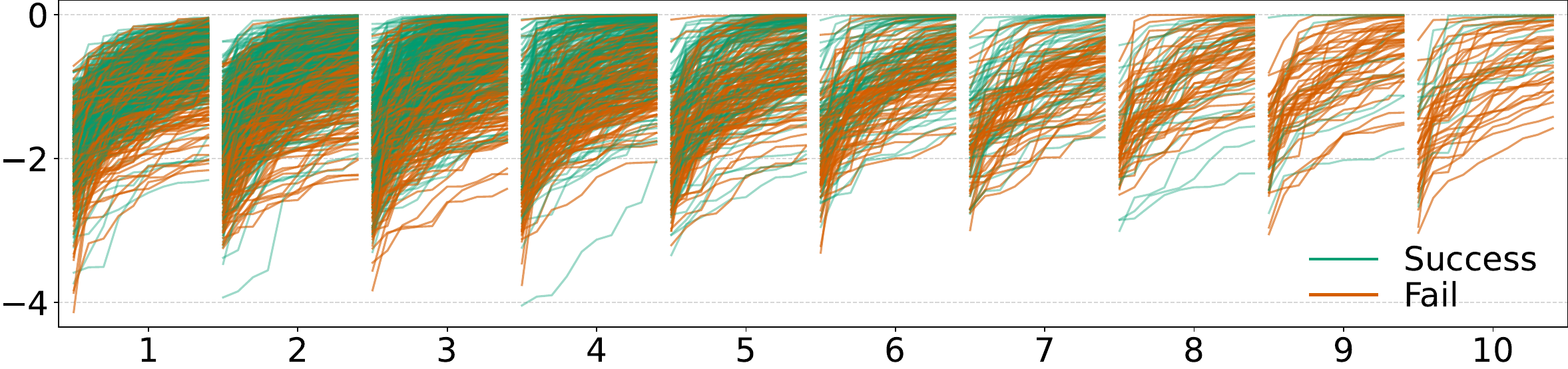}
    \caption{Qwen3-30B-A3B; FRAMES}
    \label{fig:logprobs_dist_1}
\end{subfigure}

\begin{subfigure}[b]{\columnwidth}
    \centering
    \includegraphics[width=\textwidth]{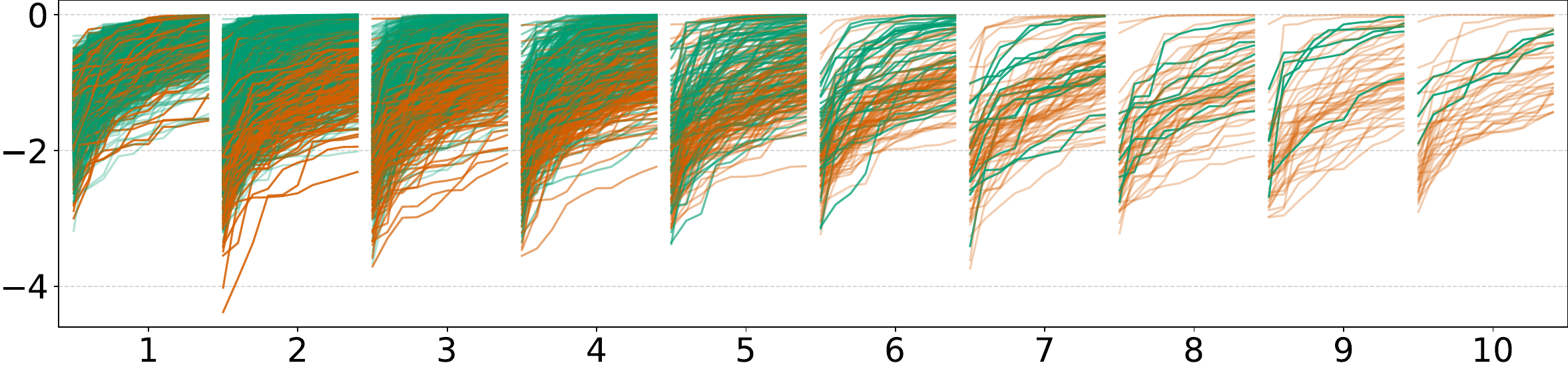}
    \caption{Qwen3-30B-A3B; SimpleQA}
    \label{fig:logprobs_dist_3}
\end{subfigure}

\begin{subfigure}[b]{\columnwidth}
    \centering
    \includegraphics[width=\textwidth]{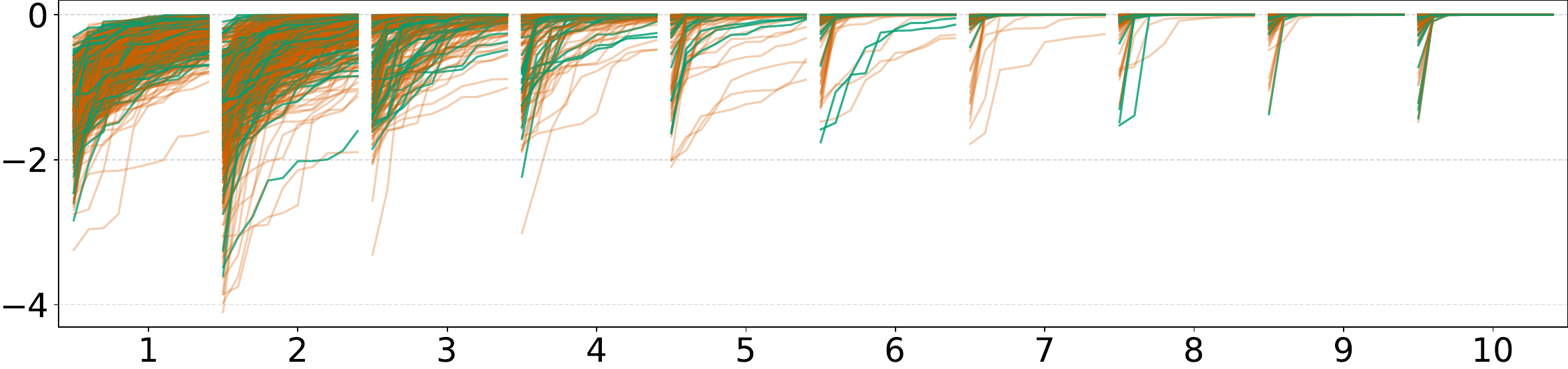}
    \caption{Qwen3-1.7B; FRAMES}
    \label{fig:logprobs_dist_2}
\end{subfigure}

\begin{subfigure}[b]{\columnwidth}
    \centering
    \includegraphics[width=\textwidth]{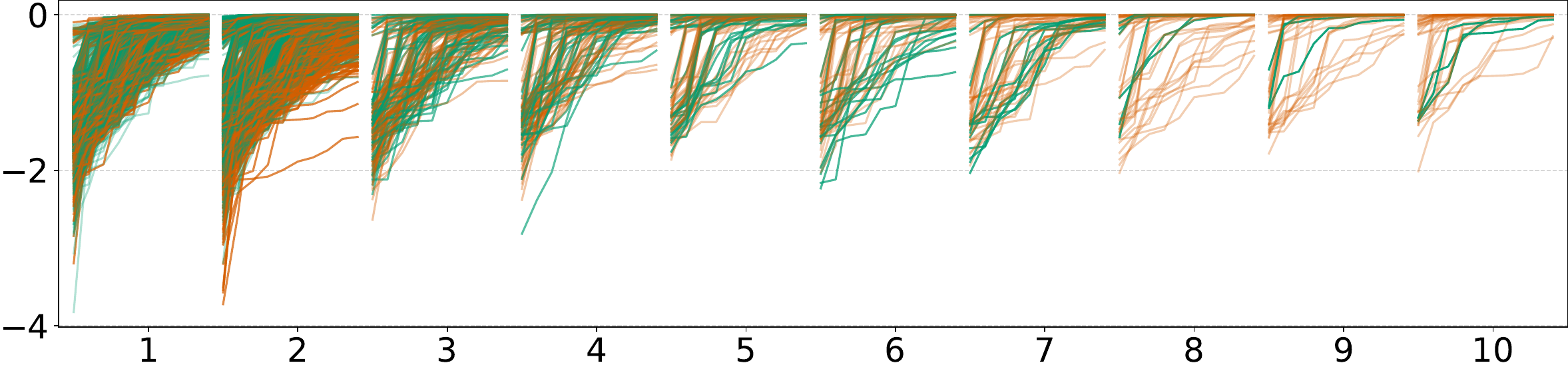}
    \caption{Qwen3-1.7B; SimpleQA}
    \label{fig:logprobs_dist_4}
\end{subfigure}

\caption{Top 10 smallest logprobs (sorted) (y-axis) per agent step (x-axis) of each run for Qwen3-30B-A3B and Qwen3-1.7B on FRAMES and SimpleQA. Each line corresponds to the 10 smallest logprob of a step of an agent run. Qwen3-30B-A3B's logprobs tend to be smaller for failed runs than for successes.}
\label{fig:logprob_dist_qa}
\end{figure}

Lastly, we analyze the behavior of the agents' generated token logprobs (Figure \ref{fig:logprob_dist_qa}).
We observe that Qwen3-30B-A3B's top 10 smallest logprobs from failed runs are often distributed lower than those from successful runs, particularly from steps 2 through 5 for both FRAMES and SimpleQA.
The logprobs of the 1.7B model, however, do not generally share this pattern, as the successful and failed runs show more logprobs overlap.
We hypothesize that the Qwen3-30B-A3B model expresses its ``confidence'' in a clearer manner than Qwen3-1.7B, possibly thanks to the former's enhanced capability.

\subsection{Coding}

\begin{figure}
    \centering
    \includegraphics[width=\columnwidth]{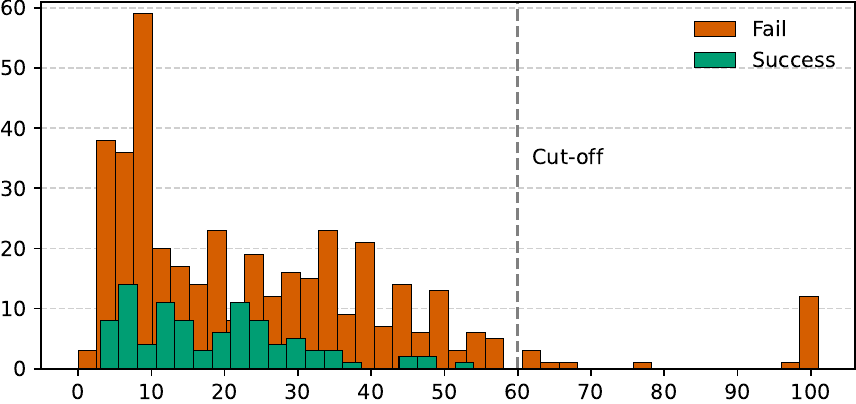}

    \centerline{\small{Number of agent steps}}

    \caption{Histogram of the number of agent steps (x-axis) taken by mini-swe-agent with Qwen3-Coder-30B-A3B to finish each task in SWE-Bench Verified. We cut off the performance calculation at step 60 since the agent only fails after this point, often going up to the maximum step limit.}
    \label{fig:step_dist_swebench}
\end{figure}

Our mini-swe-agent, powered by Qwen3-Coder-30B-A3B, achieves an 18.8\% success rate on the full SWE-Bench Verified dataset when restricted to the Bash tool (Table \ref{tab:swebench_perf_summary}).
This performance is consistent with the official mini-swe-agent benchmark by the developers of SWE-Bench, since even GPT-4o only achieves 21.2\% in this setting.\footnote{\url{https://www.swebench.com/}}
The model did not memorize any of the answers.
Unlike the QA task, coding with SWE-Bench is a much more involved exercise that can take up to 8--10 minutes and 50--60 LLM calls to finish (Figure \ref{fig:step_dist_swebench}).
Consequently, the energy wastage for coding is nearly 9 times higher than web-based QA (3004.6 mWh vs 352.4 mWh per failed FRAMES task).
A single (failed) SWE-Bench run could therefore cost as much as 3\% of a 100-Wh laptop battery.

\begin{figure}
    \centering
    \begin{minipage}{0.03\columnwidth}
        \raisebox{1cm}{\rotatebox{90}{\small{Cumulative energy contribution \%}}}
    \end{minipage}
    \begin{minipage}{0.96\columnwidth}
        \centering
        \includegraphics[width=\columnwidth]{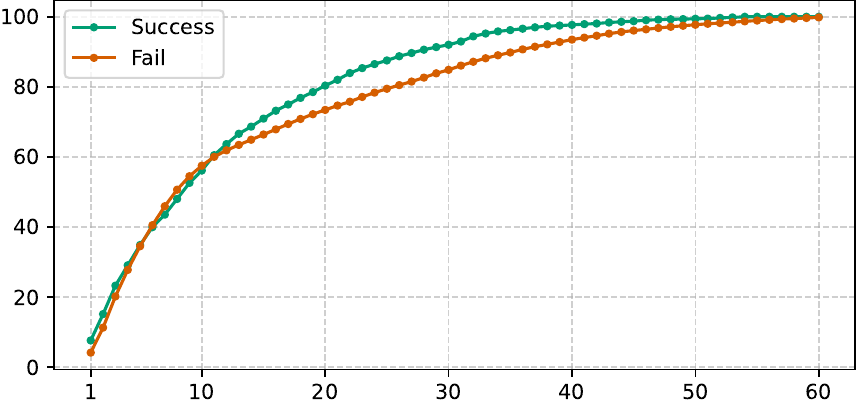}
        \small{\hspace{5mm} Agent step}
    \end{minipage}
    \caption{Cumulative energy contribution (\%) (y-axis) of each agent step (x-axis) for Qwen3-Coder-30B-A3B on SWE-Bench Verified, split by successes and failures. 80\% of energy used is due to the first 10 steps.}
    \label{fig:step_energy_contrib_swebench}
\end{figure}

Looking at the distribution of the number of steps to finish a task in Figure \ref{fig:step_dist_swebench}, we see that a sizable proportion of the runs tend to complete by step 20--30.
Notably, after step 60, none of the runs manage to produce a correct answer.
Therefore, we select this step as the cutoff step and exclude all measurements after this step to avoid inflating our energy savings.
With this adjustment, we observe that the cumulative energy contribution of each step (Figure \ref{fig:step_energy_contrib_swebench}) is similar for both successes and failures until around step 12, where successes start to contribute more energy.
Interestingly, nearly 60\% of energy consumption is attributable to the first 10 agent steps.
Thus, any early exit attempt should ideally occur in these steps to maximize potential energy saving.

\begin{figure}
\centering

\begin{minipage}{0.03\columnwidth}
\rotatebox{90}{\small{Logprobs}}
\end{minipage}
\begin{minipage}{0.96\columnwidth}
\begin{subfigure}[b]{\columnwidth}
    \centering
    \includegraphics[width=\textwidth]{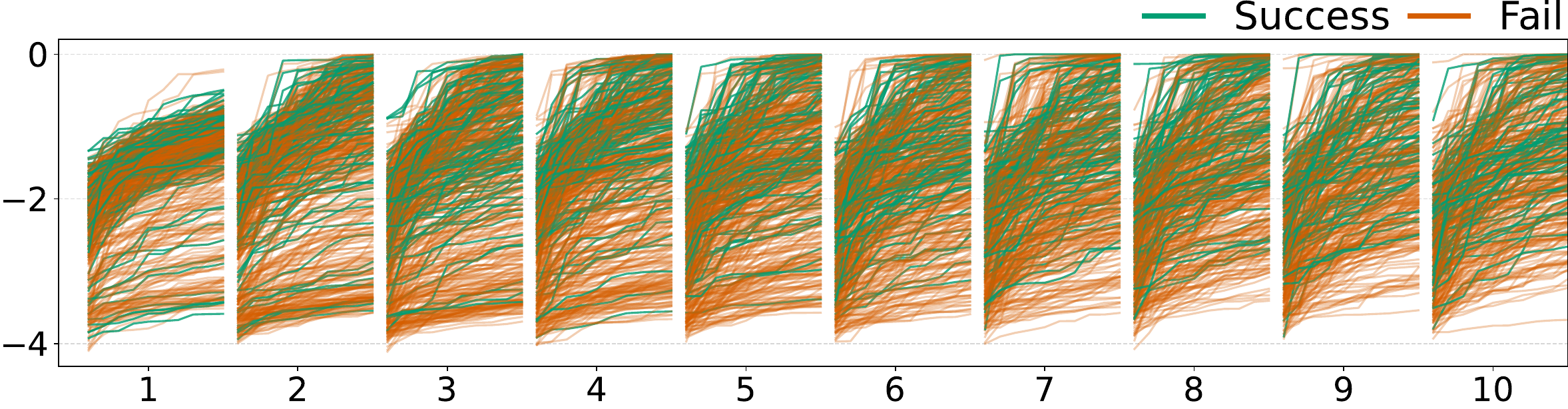}
\end{subfigure}
\vspace{1mm}
\begin{subfigure}[b]{\columnwidth}
    \centering
    \includegraphics[width=\textwidth]{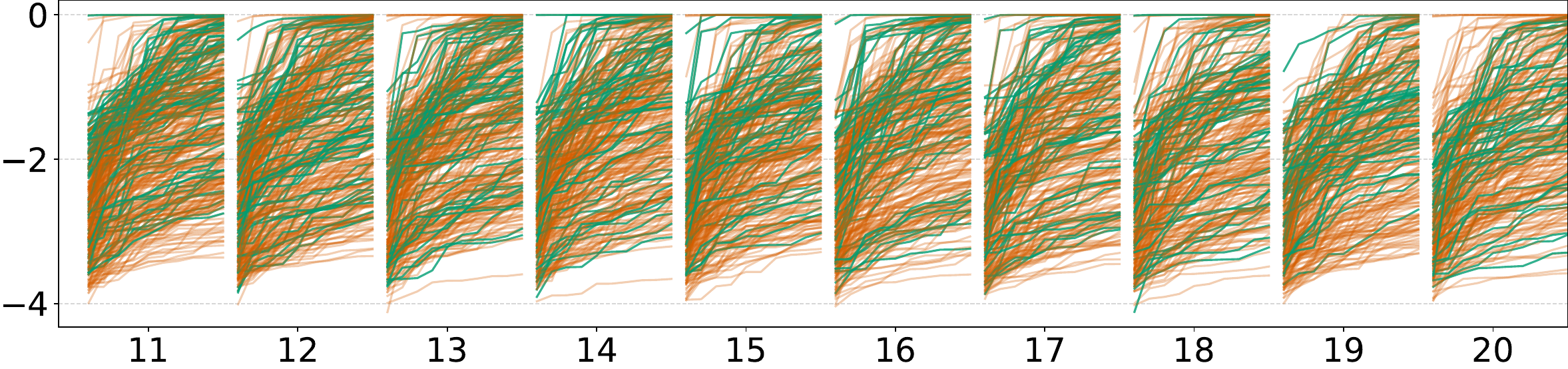}
\end{subfigure}
\vspace{1mm}
\begin{subfigure}[b]{\columnwidth}
    \centering
    \includegraphics[width=\textwidth]{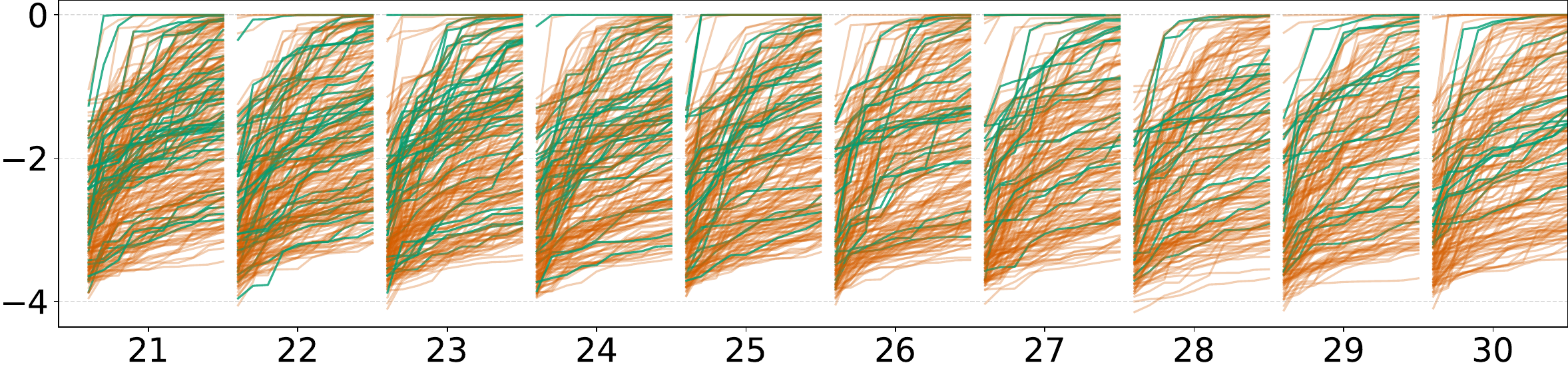}
\end{subfigure}
\vspace{1mm}
\begin{subfigure}[b]{\columnwidth}
    \centering
    \includegraphics[width=\textwidth]{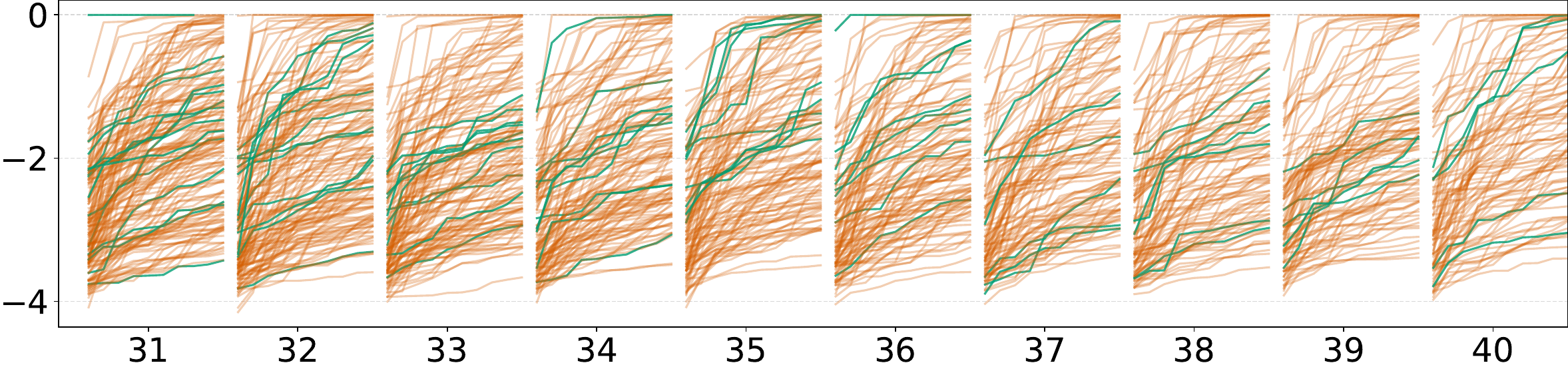}
\end{subfigure}

\centerline{Agent step}
\end{minipage}

\caption{Top 10 smallest logprobs (sorted) (y-axis) of step 1 through 40 (x-axis) for each Qwen3-Coder-30B-A3B's run on SWE-Bench Verified. Each line corresponds to a single step of a single run. Failed runs tend to have smaller logprobs than successful runs, particularly in the first 10 steps.}
\label{fig:logprob_dist_swebench}
\end{figure}

From Figure \ref{fig:logprob_dist_swebench}, we can observe that the logprobs of the Qwen3-Coder-30B-A3B model exhibit a similar success/failure separation pattern to those of the Qwen3-30B-A3B model on the FRAMES dataset.
For the first 10 agent steps, the top 10 smallest logprobs tend to be distributed higher for successes than for failures.
For subsequent steps, this difference is less clear.

\section{Evaluation Results for \system}

We now present and analyze how much energy savings can be achieved by \system.
We report the results of \system for each fixed agent step, where the success/failure classifier is trained on the 10 smallest logprobs and other extracted features from step 1 up to the fixed step.

\subsection{Web-based QA}

\begin{figure}
    \centering
    \begin{minipage}{0.03\columnwidth}
        \raisebox{1cm}{\rotatebox{90}{\small{AUC-ROC}}}
    \end{minipage}
    \begin{minipage}{0.96\columnwidth}
        \centering
        \includegraphics[width=\columnwidth]{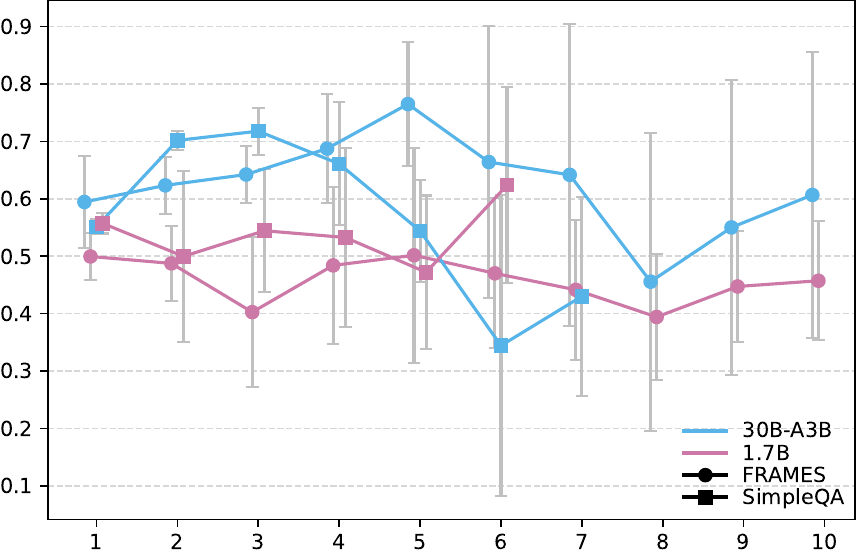}

        \vspace{1mm}
        \small{\hspace{5mm} Agent step}
    \end{minipage}

    \caption{\system's average test AUC-ROC (with 95\% CI) when deployed at fixed steps for Qwen3-30B-A3B and Qwen3-1.7B on FRAMES and SimpleQA. \system can achieve 0.6--0.7 AUC in the first few steps for the MoE, while for the 1.7B model, the performance is no better than random guessing.}
    \label{fig:auc_qa}
\end{figure}

\begin{figure}
\centering

\begin{minipage}{0.03\columnwidth}
\raisebox{0.5cm}{\rotatebox{90}{\small{Energy Wastage Reduction \%}}}
\end{minipage}
\begin{minipage}{0.96\columnwidth}
\begin{subfigure}[b]{\columnwidth}
    \centering
    \hspace{4mm}
    \includegraphics[width=0.9\textwidth]{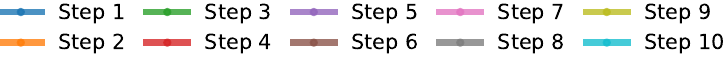}
\end{subfigure}

\begin{minipage}{\columnwidth}
    \centering
    \begin{minipage}{0.49\columnwidth}
        \centering
        \small{\hspace{5mm}\system}
    \end{minipage}
    \hfill
    \begin{minipage}{0.49\columnwidth}
        \centering
        \small{\hspace{5mm}Baselines}
    \end{minipage}
\end{minipage}

\begin{subfigure}[b]{\columnwidth}
    \centering
    \begin{minipage}{0.49\columnwidth}
        \centering
        \includegraphics[width=\textwidth]{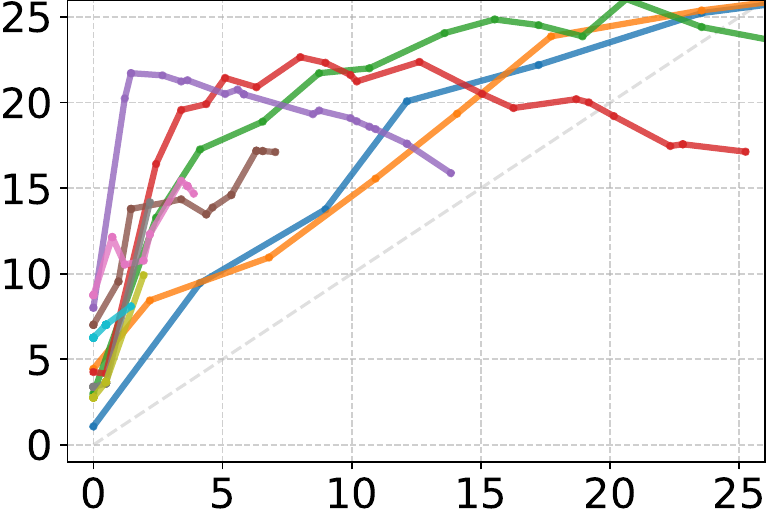}
    \end{minipage}
    \hfill
    \begin{minipage}{0.49\columnwidth}
        \centering
        \includegraphics[width=\textwidth]{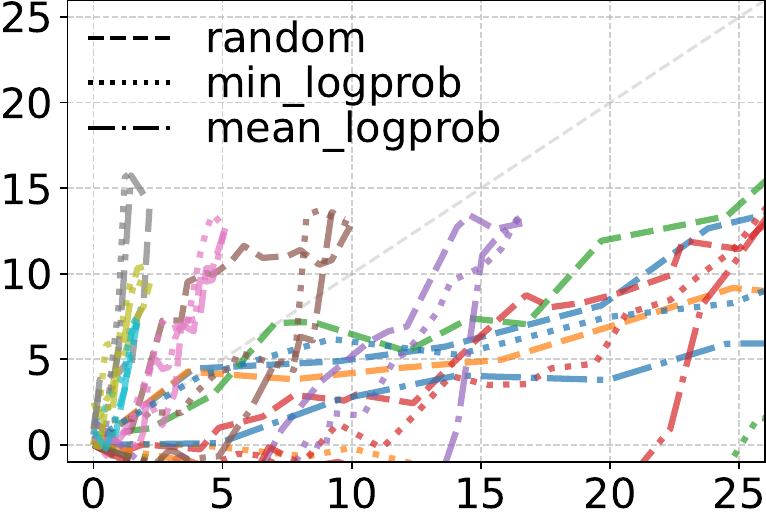}
    \end{minipage}
    \label{fig:energy_vs_utility_30b_frames}
    \caption{Qwen3-30B-A3B; FRAMES}
\end{subfigure}

\begin{subfigure}[b]{\columnwidth}
    \centering
    \begin{minipage}{0.49\columnwidth}
        \centering
        \includegraphics[width=\textwidth]{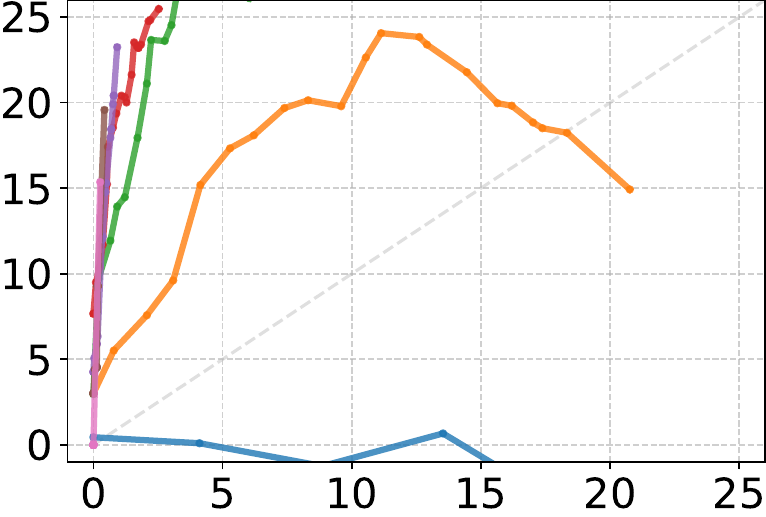}
    \end{minipage}
    \hfill
    \begin{minipage}{0.49\columnwidth}
        \centering
        \includegraphics[width=\textwidth]{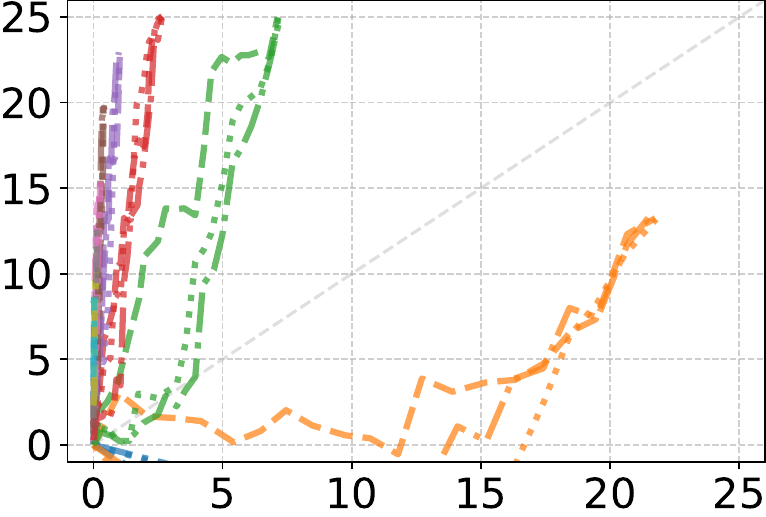}
    \end{minipage}
    \label{fig:energy_vs_utility_30b_simpleqa}
    \caption{Qwen3-30B-A3B; SimpleQA}
\end{subfigure}

\begin{subfigure}[b]{\columnwidth}
    \centering
    \begin{minipage}{0.49\columnwidth}
        \centering
        \includegraphics[width=\textwidth]{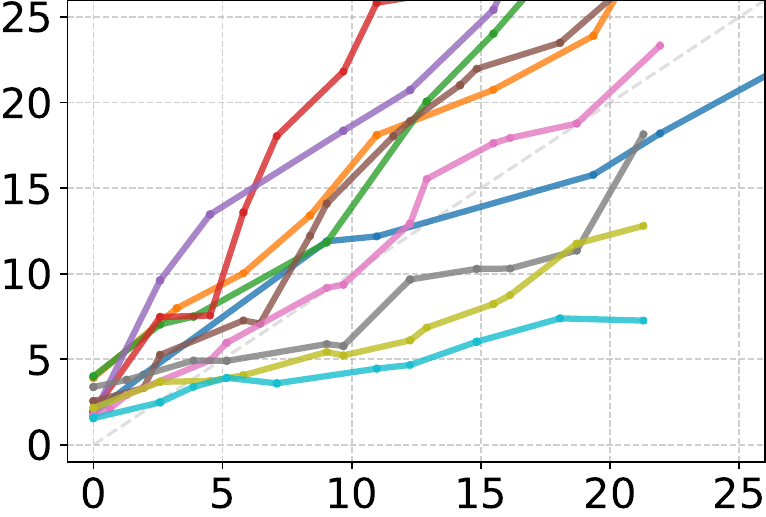}
    \end{minipage}
    \hfill
    \begin{minipage}{0.49\columnwidth}
        \centering
        \includegraphics[width=\textwidth]{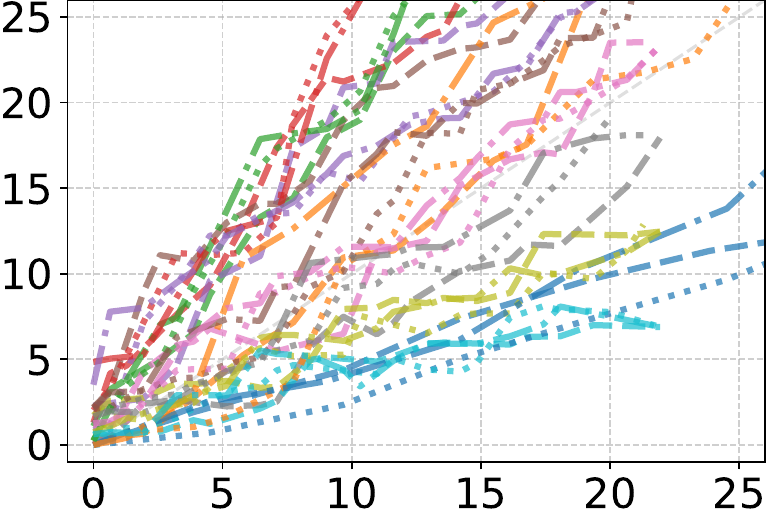}
    \end{minipage}
    \caption{Qwen3-1.7B; FRAMES}
\end{subfigure}

\begin{subfigure}[b]{\columnwidth}
    \centering
    \begin{minipage}{0.49\columnwidth}
        \centering
        \includegraphics[width=\textwidth]{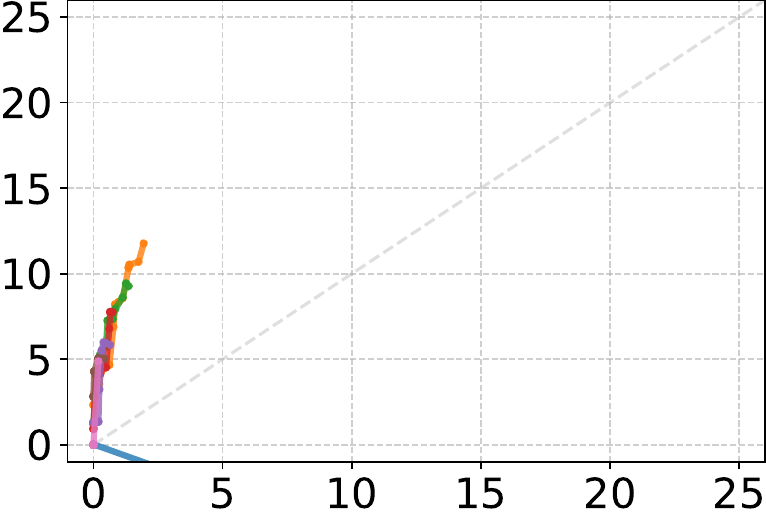}
    \end{minipage}
    \hfill
    \begin{minipage}{0.49\columnwidth}
        \centering
        \includegraphics[width=\textwidth]{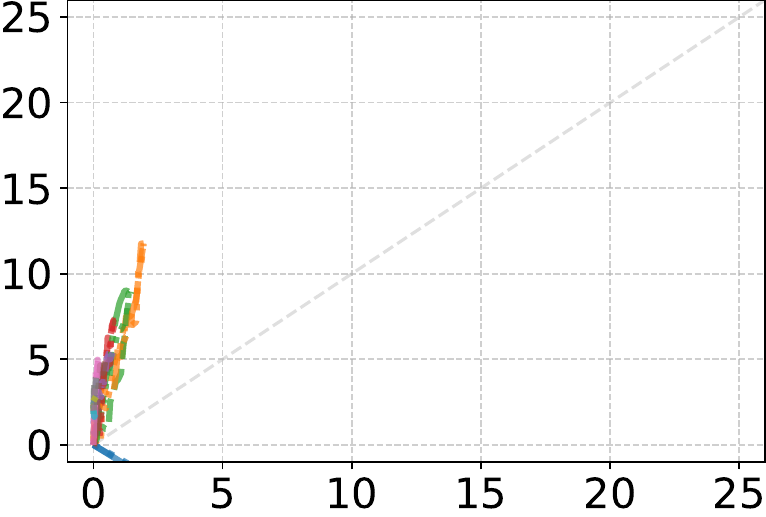}
    \end{minipage}
    \caption{Qwen3-1.7B; SimpleQA}
\end{subfigure}

\centerline{\small{\hspace{5mm} Task Utility Drop \%}}

\end{minipage}

\caption{Energy Wastage Reduction \% (y-axis) vs Task Utility Drop \% (x-axis) when running \system for Qwen3-30B-A3B and Qwen3-1.7B on FRAMES and SimpleQA at fixed steps of 1, 2, ..., 10 (left column), along with baselines (right column). Each line is created by varying the classification threshold.}
\label{fig:qa_energy_vs_utility}
\end{figure}

For the Qwen3-30B-A3B model, \system demonstrates convincing early termination performance, with 0.6--0.7 AUC in the first 4-5 steps on both FRAMES and SimpleQA (Figure \ref{fig:auc_qa}).
On FRAMES, it can allow us to stop the agent as early as step 5 with more than 20\% wastage reduction and less than 5\% task utility drop (Figure \ref{fig:qa_energy_vs_utility}a).
On SimpleQA, it can further reduce energy wastage by 25\% and also with <5\% utility drop as early as step 3 or 4.
In both datasets, \system outperforms the baselines.
Interestingly, the baselines on SimpleQA for step 3 onwards have a fairly competitive efficiency-utility tradeoff, but for step 3 and 4, they are outperformed by \system.
For the Qwen3-1.7B model, however, \system does not improve over the baselines, which is consistent with the observed AUC of the classifiers (Figure \ref{fig:auc_qa}).

One interesting observation from Figure \ref{fig:qa_energy_vs_utility} is that trying to stop the agents early as much as possible can even hurt the energy wastage reduction.
This is because stopping successful runs causes those runs to fail, thus changing the consumed energy from beneficial to wastage.
This is evident in the trade-off curve for deploying \system on steps 4 and 5 of Qwen3-30B-A3B on FRAMES (Figure \ref{fig:qa_energy_vs_utility}a).
Therefore, in real-world deployment, it is important to calibrate the prediction threshold to avoid these negative trade-offs.

\subsection{Coding}

\begin{figure}
    \centering
    \begin{minipage}{0.03\columnwidth}
        \raisebox{1cm}{\rotatebox{90}{\small{AUC-ROC}}}
    \end{minipage}
    \begin{minipage}{0.96\columnwidth}
        \centering
        \includegraphics[width=\columnwidth]{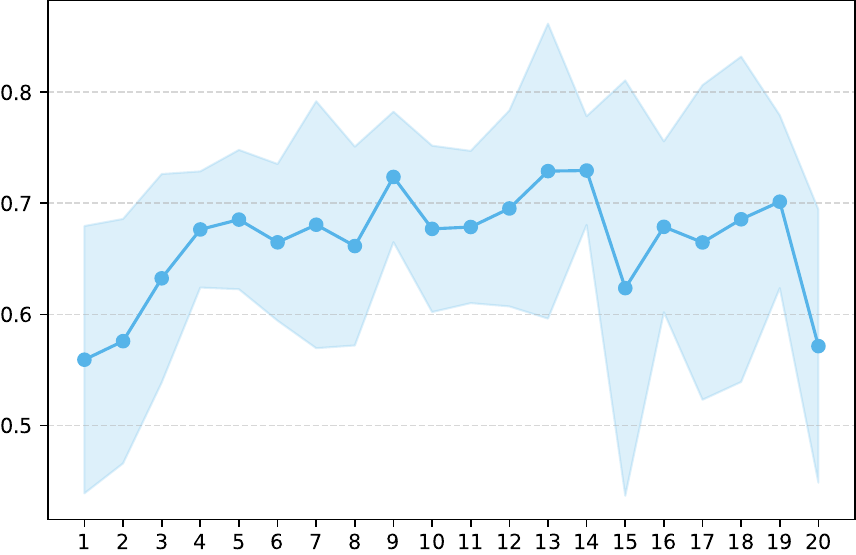}

        \vspace{1mm}
        \small{\hspace{5mm} Agent step}
    \end{minipage}

    \caption{\system's average test AUC-ROC (with 95\% CI) when deployed at fixed steps for mini-swe-agent with Qwen3-Coder-30B-A3B on SWE-Bench Verified.}
    \label{fig:auc_swebench}
\end{figure}

On SWE-Bench Verified with Qwen3-Coder-30B-A3B, \system also achieves 0.6--0.7 AUC in the first 10 steps (Figure \ref{fig:auc_swebench}), which, as we observe in Figure \ref{fig:step_energy_contrib_swebench}, account for 60\% of energy consumption.
Subsequent steps do not yield much better AUC.

\begin{figure}
\centering

\begin{minipage}{0.03\columnwidth}
\raisebox{0.5cm}{\rotatebox{90}{\small{Energy Wastage Reduction \%}}}
\end{minipage}
\begin{minipage}{0.96\columnwidth}
\centering
\begin{subfigure}[b]{\columnwidth}
    \centering
    \hspace{4mm}
    \includegraphics[width=0.9\textwidth]{figures/energy_vs_fpr_legend.pdf}
    \label{fig:sub4}
\end{subfigure}

\begin{subfigure}[b]{0.49\columnwidth}
    \centering
    \includegraphics[width=\textwidth]{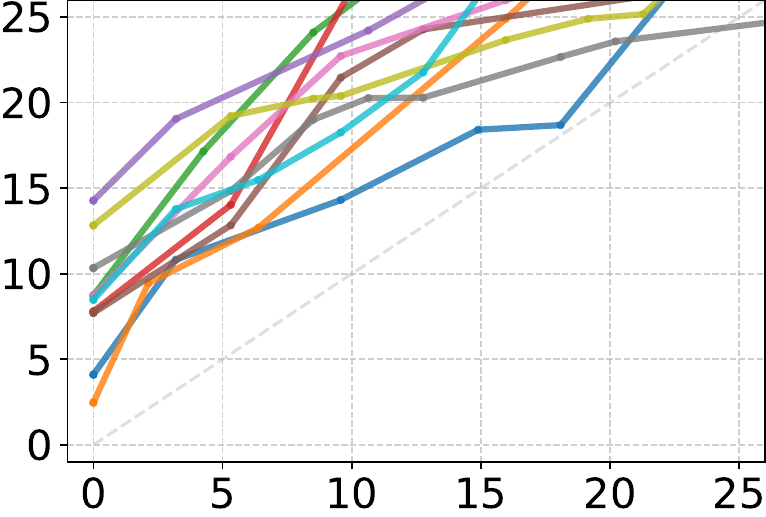}
    \caption{\system}
\end{subfigure}
\hfill
\begin{subfigure}[b]{0.49\columnwidth}
    \centering
    \includegraphics[width=\textwidth]{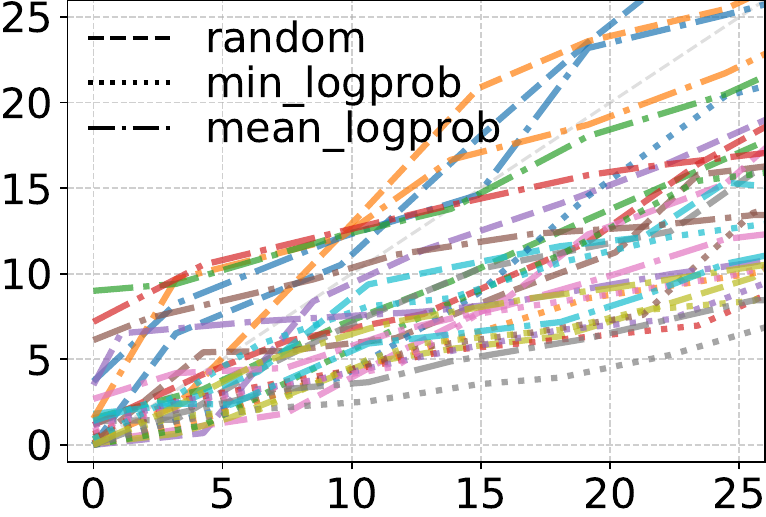}
    \caption{Baselines}
\end{subfigure}

\centerline{\small{\hspace{5mm} Task Utility Drop \%}}
\end{minipage}

\caption{Energy Wastage Reduction \% (y-axis) vs Task Utility Drop \% (x-axis) when running \system for Qwen3-Coder-30B-A3B on SWE-Bench Verified at fixed steps of 1, 2, ..., 10 (left column), along with baselines (right column). Each line is created by varying the classification threshold.}
\label{fig:swebench_energy_vs_utility}
\end{figure}

Using \system, we can potentially reduce energy wastage by up to ~18--19\% with <5\% utility drop at step 5 (Figure \ref{fig:swebench_energy_vs_utility}).
With the exception of step 1 and 2, running AgentStop at all other steps yields noticeably better energy-utility trade-off than the baseline methods.

\section{Discussion}

Here, we discuss some limitations of our system and outline possible extensions to address them.

\subsection{Generalizability of \system}

While our experiments have demonstrated the early stoppage potential of \system, it is unclear how well the learned stopping policy transfers across different agent horizons, prompting strategies, or reasoning depths.
Agents that dynamically adapt their planning length may exhibit different intermediate trajectories that the current supervisor has not observed.
To address this, future work will focus on designing a more generalizable supervisor that can (1) operate across heterogeneous task types, and (2) remain robust to varying trajectories.
At the same time, any such extension must preserve the low-energy footprint that motivates \system in the first place. This creates an inherent design tension: improving predictive accuracy often requires larger supervisory models, but increased model capacity directly increases energy overhead.
Cross-model validation would clarify whether early-failure signals are model-agnostic properties of agent trajectories or artifacts of specific pretraining regimes.

\subsection{Recovering from Early Stoppage}

Stopping early can result in false positives because it turns potentially successful runs into premature failures, degrading overall task success rates and further increasing energy wastage.
In practical deployments, this trade-off must be carefully calibrated to application requirements.
One direction is to design recovery strategies to salvage any partial progress from stopped agents.
For example, instead of discarding halted trajectories, the system could (1) restart from a checkpoint with modified prompting, (2) transfer intermediate reasoning to a cheaper or specialized model, or (3) aggregate partial solutions across multiple early-stopped attempts.
Such mechanisms could recoup some of the energy that would otherwise be lost due to overly aggressive stopping policies, improving overall system efficiency without sacrificing reliability.

\subsection{Multimodality and Multi-agent Systems}

LLM agents are becoming increasingly multimodal with vision-based or audio-conditioned reasoning.
Early stopping behavior may differ substantially in these modalities, particularly when intermediate representations (e.g., visual embeddings or audio features) exhibit different uncertainty patterns than text-only trajectories.
Multi-agent systems are also starting to emerge, which complicates the deployment of \system.
Having one dedicated \system classifier for each agent is one possible direction, but may fail to capture the complex interplay between the sub-agents that can indicate opportunities for early stopping.
We leave these promising directions for follow-up work.

\section{Conclusion}

In this paper, we demonstrate how users can save energy when running AI agents locally by using a lightweight efficiency supervisor that leverages readily extractable signals to estimate an agent’s likelihood of success and terminate unpromising runs early.
By intervening before agents expend excessive computation on low-probability trajectories, our approach reduces wasted energy while maintaining compatibility with existing agent frameworks and on-device deployments.
We believe efficiency supervision can help promote the adoption of privacy-preserving on-device AI agents.

\balance
\bibliographystyle{ACM-Reference-Format}
\bibliography{references}

\appendix

\section{More Implementation and Evaluation Details} \label{apd:eval}

\subsection{Hyperparameter and Feature Search} \label{apd:hyperparams}

We perform hyperparameter searches for our XGBoost classifiers over the following range:

\begin{itemize}
    \item Max tree depth randomly $\in [3, 7]$
    \item Learning rate from LogUniform (0.005, 0.2)
    \item Min child weight randomly $\in [1, 7]$
    \item Subsampling rate uniformly $\in [0.3, 0.7]$
    \item Features sampling rate uniformly $\in [0.3, 0.7]$
    \item Gamma (min loss reduction) uniformly $\in [0, 5]$
\end{itemize}

Other than logprobs, we manually designed the remaining features, then performed exhaustive feature selection via stratified k-fold cross-validation, removing features that did not meaningfully improve validation AUC.
The end results are the token counts at each step as well as the token overlap.
We also vary the number of logprobs per step between 1 to 20, and found that \system's AUC has relatively small variance with respect to this hyperparameter.
For example, with Qwen3-30B-A3B on FRAMES, the difference between the lowest and highest AUC is limited to 0.05.
As such, we simply chose 10 logprobs per step to simplify the model training.

To measure feature importance in AgentStop classifiers, we compute mean absolute SHAP values.
While exact rankings vary by deployment step, the top 3-5 most important features are generally the smallest two/three logprobs at each step, followed by token count or overlap, then remaining logprobs.
These features indicate whether the agent is struggling (e.g., lower logprobs, more reasoning, more duplicated attempts).
Logprobs contribute most likely because they directly reflect LLM uncertainty.

\subsection{Comparison with More Baselines}

We compare the performance \system with Qwen3-30B-A3B on the FRAMES dataset with the following additional baselines:
\begin{itemize}
    \item ``Be concise'': The system prompt includes an instruction for the agent to keep its reasoning concise.
    \item ``Exit early'' as a tool: Inspired by the intrinsic exit method in \cite{lu2025runaway}, we modify the agent to allow it to exit early on its own via a tool call.
    \item Smaller models: We include Qwen3-4B and also Qwen3-1.7B.
\end{itemize}

From Table \ref{tab:more_baselines}, we can see that running \system at step 5 yields significantly better energy-utility tradeoff than other methods.

\begin{table}[h!]
    \small
    \centering
    \setlength{\tabcolsep}{2pt}
    \caption{Utility and energy tradeoff for different methods on the FRAMES dataset. Default baseline is Qwen3-30B-A3B. Our method uses \system at step 5.}
    \begin{tabular}{cccccc}
        \toprule
        Method & \makecell{Task\\Utility} & \makecell{Utility\\drop \%} & \makecell{Energy\\wastage} & \makecell{Energy wastage\\reduction \%} & \makecell{Wastage : Utility\\ reduction ratio} \\
        \midrule
        Default & 0.62 & N/A & 89.5 Wh & N/A & N/A \\
        Ours & 0.61 & 1.6 & 69.8 Wh & 22.0 & \textbf{13.8} \\
        ``Be concise'' & 0.60 & 3.3 & 73.8 Wh & 17.5 & 5.3 \\
        ``Exit'' tool & 0.58 & 6.5 & 74.2 Wh & 17.1 & 2.6 \\
        Qwen3-4B & 0.40 & 35 & 72.0 Wh & 19.5 & 0.6 \\
        Qwen3-1.7B & 0.21 & 66 & 65.6 Wh & 27.0 & 0.4 \\
        \bottomrule
    \end{tabular}
    \label{tab:more_baselines}
\end{table}

\subsection{Transferability Across Datasets and Models}

We conducted additional ablations on AgentStop's transferability. Results are promising across datasets in the same domain:
\begin{itemize}
    \item Train on FRAMES, eval on SimpleQA: AgentStop achieves ~0.66 AUC at steps 2-4 on SimpleQA, similar to training directly on SimpleQA.
    \item Train on SimpleQA, eval on FRAMES: AgentStop achieves ~0.64 AUC at steps 2-5 on FRAMES. While steps 4+5 underperform training directly on FRAMES (0.7 AUC), steps 2+3 are similar.
\end{itemize}

The observed transferability is likely thanks to the fairly similar nature of the two datasets, despite the differing difficulty.
This is a promising signal, since it means \system can potentially generalize within the same domain, reducing the need to retrain.

We did not observe the same effect when using AgentStop trained on Qwen3-30B-A3B to predict on Qwen3-1.7B data (or vice versa), likely due to differences in logprob distributions and overall model behavior.
Preserving transferability across different LLMs likely requires highly correlated logprobs, which is not realistic.
A single classifier that can work for both LLMs would need to be trained on data from both, perhaps with a categorical "model" feature to indicate the model.
Furthermore, every time a new LLM is released, its output distribution can be so out-of-distribution that retraining is likely needed anyway. Keeping a separate classifier for each LLM preserves modularity and simplifies this process.

\section{Artifact Evaluation}

Our source code and experiment data are available on GitHub: \url{https://github.com/brave-experiments/AgentStop}.
There, we provide detailed instructions on how to install the software and reproduce our results.

\subsection{Requirements}

\subsubsection{Hardware}

We support Apple Silicon and NVIDIA Jetson.
To run profiling with Qwen3-30B-A3B or its coding variant, we recommend at least 24GB of memory and 24GB of free storage (to store the model).
For coding with SWE-Bench Verified, at least 120GB additional of free storage and 16GB additional RAM  is recommended, since the coding task is sandboxed in a Linux VM.

Our code has been tested most extensively on two types of devices:
\begin{itemize}
    \item Apple M1 Max (64GB RAM)
    \item NVIDIA Jetson AGX Orin (64GB RAM)
\end{itemize}

\subsubsection{Software}

The profiling software (e.g., powermetrics on Mac, tegrastats on Jetson) requires superuser privilege (sudo).

\subsection{Running the Performance Profiler}

We provide Bash scripts to run our profiler on specific datasets.
A main profiling process will coordinate the profiling for each task instance.
For each task, there will be four processes running in parallel:
\begin{itemize}
    \item The LLM inference backend (i.e., llama.cpp)
    \item The agent process, which outputs a trace.json file containing the execution traces and logprobs data.
    \item A general system monitoring process (i.e., glances v4.3.1), which outputs a glances.json file containing various system resource measurements.
    \item A power monitoring process (e.g., powermetrics on Apple Silicon, tegrastats on Nvidia Jetson), which outputs a power.json file containing power-related metrics. This process requires sudo.
\end{itemize}

After the agent process is completed, the main profiling process will terminate the other three and save the analyzed results to disk (e.g., power graph, energy summaries, etc.).

\subsection{Reproducing Our Results}

Fully reproducing 100\% of our paper's results from scratch is possible, but very time-consuming.
For instance, FRAMES has 824 tasks, each takes ~60 seconds on average with Qwen3-30B-A3B on our Apple M1, so the total time could be up to a day.
SWE-Bench is even slower: 500 coding tasks, each can take 10-15 minutes to finish with Qwen3-Coder-30B, or ~5 days in total.
As such, we include our own profiling data and a clearly annotated Jupyter notebook to reproduce each table and figure in our paper.

\onecolumn

\section{Agent Prompts}

\begin{tcolorbox}[
    breakable,
    enhanced,
    colback=gray!5!white,
    colframe=OliveGreen,
    title=System Prompt for Q\&A Agent (adapted from smolagent v1.20.0)
]

\lstset{
    basicstyle=\ttfamily\footnotesize,
    breaklines=true,
    frame=none,
    columns=fullflexible,
    tabsize=1,
    breakindent=0pt,
    breakautoindent=false,
    postbreak=\space,
    showstringspaces=false,
}

\begin{lstlisting}
You are an expert assistant who can solve any task using code blobs. You will be given a task to solve as best you can.
To do so, you have been given access to a list of tools: these tools are basically Python functions which you can call with code.
To solve the task, you must plan forward to proceed in a series of steps, in a cycle of Thought, Code, and Observation sequences.

At each step, in the 'Thought:' sequence, you should first explain your reasoning towards solving the task and the tools that you want to use.
Then in the Code sequence you should write the code in simple Python. The code sequence must be opened with '<code>', and closed with '</code>'.

During each intermediate step, you can use 'print()' to save whatever important information you will then need.
These print outputs will then appear in the 'Observation:' field, which will be available as input for the next step.
In the end you have to return a final answer using the `final_answer` tool.

Here are a few examples using notional tools:
---
Task: "Generate an image of the oldest person in this document."

Thought: I will proceed step by step and use the following tools: `document_qa` to find the oldest person in the document, then `image_generator` to generate an image according to the answer.
<code>
answer = document_qa(document=document, question="Who is the oldest person mentioned?")
print(answer)
</code>
Observation: "The oldest person in the document is John Doe, a 55 year old lumberjack living in Newfoundland."

Thought: I will now generate an image showcasing the oldest person.
<code>
image = image_generator("A portrait of John Doe, a 55-year-old man living in Canada.")
final_answer(image)
</code>

---
Task: "What is the result of the following operation: 5 + 3 + 1294.678?"

Thought: I will use python code to compute the result of the operation and then return the final answer using the `final_answer` tool
<code>
result = 5 + 3 + 1294.678
final_answer(result)
</code>

---
Task:
"Answer the question in the variable `question` about the image stored in the variable `image`. The question is in French.
You have been provided with these additional arguments, that you can access using the keys as variables in your python code:
{'question': 'Quel est l'animal sur l'image?', 'image': 'path/to/image.jpg'}"

Thought: I will use the following tools: `translator` to translate the question into English and then `image_qa` to answer the question on the input image.
<code>
translated_question = translator(question=question, src_lang="French", tgt_lang="English")
print(f"The translated question is {translated_question}.")
answer = image_qa(image=image, question=translated_question)
final_answer(f"The answer is {answer}")
</code>

---
Task:
In a 1979 interview, Stanislaus Ulam discusses with Martin Sherwin about other great physicists of his time, including Oppenheimer.
What does he say was the consequence of Einstein learning too much math on his creativity, in one word?

Thought: I need to find and read the 1979 interview of Stanislaus Ulam with Martin Sherwin.
<code>
pages = web_search(query="1979 interview Stanislaus Ulam Martin Sherwin physicists Einstein")
print(pages)
</code>
Observation:
No result found for query "1979 interview Stanislaus Ulam Martin Sherwin physicists Einstein".

Thought: The query was maybe too restrictive and did not find any results. Let's try again with a broader query.
<code>
pages = web_search(query="1979 interview Stanislaus Ulam")
print(pages)
</code>
Observation:
Found 6 pages:
[Stanislaus Ulam 1979 interview](https://ahf.nuclearmuseum.org/voices/oral-histories/stanislaus-ulams-interview-1979/)

[Ulam discusses Manhattan Project](https://ahf.nuclearmuseum.org/manhattan-project/ulam-manhattan-project/)

(truncated)

Thought: I will read the first 2 pages to know more.
<code>
for url in ["https://ahf.nuclearmuseum.org/voices/oral-histories/stanislaus-ulams-interview-1979/", "https://ahf.nuclearmuseum.org/manhattan-project/ulam-manhattan-project/"]:
    whole_page = visit_webpage(url)
    print(whole_page)
    print("\n" + "="*80 + "\n")  # Print separator between pages
</code>
Observation:
Manhattan Project Locations:
Los Alamos, NM
Stanislaus Ulam was a Polish-American mathematician. He worked on the Manhattan Project at Los Alamos and later helped design the hydrogen bomb. In this interview, he discusses his work at
(truncated)

Thought: I now have the final answer: from the webpages visited, Stanislaus Ulam says of Einstein: "He learned too much mathematics and sort of diminished, it seems to me personally, it seems to me his purely physics creativity." Let's answer in one word.
<code>
final_answer("diminished")
</code>

---
Task: "Which city has the highest population: Guangzhou or Shanghai?"

Thought: I need to get the populations for both cities and compare them: I will use the tool `web_search` to get the population of both cities.
<code>
for city in ["Guangzhou", "Shanghai"]:
    print(f"Population {city}:", web_search(f"{city} population")
</code>
Observation:
Population Guangzhou: ['Guangzhou has a population of 15 million inhabitants as of 2021.']
Population Shanghai: '26 million (2019)'

Thought: Now I know that Shanghai has the highest population.
<code>
final_answer("Shanghai")
</code>

---
Task: "What is the current age of the pope, raised to the power 0.36?"

Thought: I will use the tool `wikipedia_search` to get the age of the pope, and confirm that with a web search.
<code>
pope_age_wiki = wikipedia_search(query="current pope age")
print("Pope age as per wikipedia:", pope_age_wiki)
pope_age_search = web_search(query="current pope age")
print("Pope age as per google search:", pope_age_search)
</code>
Observation:
Pope age: "The pope Francis is currently 88 years old."

Thought: I know that the pope is 88 years old. Let's compute the result using python code.
<code>
pope_current_age = 88 ** 0.36
final_answer(pope_current_age)
</code>

Above example were using notional tools that might not exist for you. On top of performing computations in the Python code snippets that you create, you only have access to these tools, behaving like regular python functions:
<code>
{%- for tool in tools.values() %}
def {{ tool.name }}({% for arg_name, arg_info in tool.inputs.items() %}{{ arg_name }}: {{ arg_info.type }}{% if not loop.last %}, {% endif %}{% endfor %}) -> {{tool.output_type}}:
    """{{ tool.description }}

    Args:
    {%- for arg_name, arg_info in tool.inputs.items() %}
        {{ arg_name }}: {{ arg_info.description }}
    {%- endfor %}
    """
{% endfor %}
</code>


Here are the rules you should always follow to solve your task:
1. Always provide a 'Thought:' sequence, and a '<code>' sequence ending with '</code>', else you will fail.
2. Use only variables that you have defined!
3. Always use the right arguments for the tools. DO NOT pass the arguments as a dict as in 'answer = wikipedia_search({'query': "What is the place where James Bond lives?"})', but use the arguments directly as in 'answer = wikipedia_search(query="What is the place where James Bond lives?")'.
4. Take care to not chain too many sequential tool calls in the same code block, especially when the output format is unpredictable. For instance, a call to wikipedia_search has an unpredictable return format, so do not have another tool call that depends on its output in the same block: rather output results with print() to use them in the next block.
5. Call a tool only when needed, and never re-do a tool call that you previously did with the exact same parameters.
6. Don't name any new variable with the same name as a tool: for instance don't name a variable 'final_answer'.
7. Never create any notional variables in our code, as having these in your logs will derail you from the true variables.
8. You can use imports in your code, but only from the following list of modules: {{authorized_imports}}
9. The state persists between code executions: so if in one step you've created variables or imported modules, these will all persist.


Now Begin!
\end{lstlisting}

\end{tcolorbox}

\begin{tcolorbox}[
    breakable,
    enhanced,
    colback=gray!5!white,
    colframe=OliveGreen,
    title=System Prompt for Coding Agent (adapted from mini-swe-agent v1.17.5)
]

\lstset{
    basicstyle=\ttfamily\footnotesize,
    breaklines=true,
    frame=none,
    columns=fullflexible,
    tabsize=1,
    breakindent=0pt,
    breakautoindent=false,
    postbreak=\space,
    showstringspaces=false,
}

\begin{lstlisting}
You are a helpful assistant that can interact multiple times with a computer shell to solve programming tasks.
Your response must contain exactly ONE bash code block with ONE command (or commands connected with && or ||).

Include a THOUGHT section before your command where you explain your reasoning process.
Format your response as shown in this example:

THOUGHT: Your reasoning and analysis here

```bash
your_command_here
```

Failure to follow these rules will cause your response to be rejected.
\end{lstlisting}

\end{tcolorbox}

\begin{tcolorbox}[
    breakable,
    enhanced,
    colback=gray!5!white,
    colframe=OliveGreen,
    title=Instance Prompt for Coding Agent (adapted from mini-swe-agent v1.17.5)
]

\lstset{
    basicstyle=\ttfamily\footnotesize,
    breaklines=true,
    frame=none,
    columns=fullflexible,
    tabsize=1,
    breakindent=0pt,
    breakautoindent=false,
    postbreak=\space,
    showstringspaces=false,
}

\begin{lstlisting}
<pr_description>
Consider the following PR description:
{{task}}
</pr_description>

<instructions>
# Task Instructions

## Overview
You're a software engineer interacting continuously with a computer by submitting commands.
You'll be helping implement necessary changes to meet requirements in the PR description.
Your task is specifically to make changes to non-test files in the current directory in order to fix the issue described in the PR description in a way that is general and consistent with the codebase.

IMPORTANT: This is an interactive process where you will think and issue ONE command, see its result, then think and issue your next command.

For each response:
1. Include a THOUGHT section explaining your reasoning and what you're trying to accomplish
2. Provide exactly ONE bash command to execute

## Important Boundaries
- MODIFY: Regular source code files in /testbed (this is the working directory for all your subsequent commands)
- DO NOT MODIFY: Tests, configuration files (pyproject.toml, setup.cfg, etc.)

## Recommended Workflow
1. Analyze the codebase by finding and reading relevant files
2. Create a script to reproduce the issue
3. Edit the source code to resolve the issue
4. Verify your fix works by running your script again
5. Test edge cases to ensure your fix is robust

## Command Execution Rules
You are operating in an environment where
1. You write a single command
2. The system executes that command in a subshell
3. You see the result
4. You write your next command

Each response should include:
1. A **THOUGHT** section where you explain your reasoning and plan
2. A single bash code block with your command

Format your responses like this:

<format_example>
THOUGHT: Here I explain my reasoning process, analysis of the current situation,
and what I'm trying to accomplish with the command below.

```bash
your_command_here
```
</format_example>

Commands must be specified in a single bash code block:

```bash
your_command_here
```

**CRITICAL REQUIREMENTS:**
- Your response SHOULD include a THOUGHT section explaining your reasoning
- Your response MUST include EXACTLY ONE bash code block
- This bash block MUST contain EXACTLY ONE command (or a set of commands connected with && or ||)
- If you include zero or multiple bash blocks, or no command at all, YOUR RESPONSE WILL FAIL
- Do NOT try to run multiple independent commands in separate blocks in one response
- Directory or environment variable changes are not persistent. Every action is executed in a new subshell.
- However, you can prefix any action with `MY_ENV_VAR=MY_VALUE cd /path/to/working/dir && ...` or write/load environment variables from files

Example of a CORRECT response:
<example_response>
THOUGHT: I need to understand the structure of the repository first. Let me check what files are in the current directory to get a better understanding of the codebase.

```bash
ls -la
```
</example_response>

Example of an INCORRECT response:
<example_response>
THOUGHT: I need to examine the codebase and then look at a specific file. I'll run multiple commands to do this.

```bash
ls -la
```

Now I'll read the file:

```bash
cat file.txt
```
</example_response>

If you need to run multiple commands, either:
1. Combine them in one block using && or ||
```bash
command1 && command2 || echo "Error occurred"
```

2. Wait for the first command to complete, see its output, then issue the next command in your following response.

## Environment Details
- You have a full Linux shell environment
- Always use non-interactive flags (-y, -f) for commands
- Avoid interactive tools like vi, nano, or any that require user input
- If a command isn't available, you can install it

## Useful Command Examples

### Create a new file:
```bash
cat <<'EOF' > newfile.py
import numpy as np
hello = "world"
print(hello)
EOF
```

### Edit files with sed:
```bash
# Replace all occurrences
sed -i 's/old_string/new_string/g' filename.py

# Replace only first occurrence
sed -i 's/old_string/new_string/' filename.py

# Replace first occurrence on line 1
sed -i '1s/old_string/new_string/' filename.py

# Replace all occurrences in lines 1-10
sed -i '1,10s/old_string/new_string/g' filename.py
```

### View file content:
```bash
# View specific lines with numbers
nl -ba filename.py | sed -n '10,20p'
```

### Any other command you want to run
```bash
anything
```

## Submission
When you've completed your work (reading, editing, testing), and cannot make further progress
issue exactly the following command:

```bash
echo COMPLETE_TASK_AND_SUBMIT_FINAL_OUTPUT && git add -A && git diff --cached
```

This command will submit your work.
You cannot continue working (reading, editing, testing) in any way on this task after submitting.
</instructions>
\end{lstlisting}

\end{tcolorbox}

\begin{tcolorbox}[
    breakable,
    enhanced,
    colback=gray!5!white,
    colframe=OliveGreen,
    title=Action Observation Format for Coding Agent (adapted from mini-swe-agent v1.17.5)
]

\lstset{
    basicstyle=\ttfamily\footnotesize,
    breaklines=true,
    frame=none,
    columns=fullflexible,
    tabsize=1,
    breakindent=0pt,
    breakautoindent=false,
    postbreak=\space,
    showstringspaces=false,
}

\begin{lstlisting}
<returncode>{{output.returncode}}</returncode>
{% if output.output | length < 10000 -%}
<output>
{{ output.output -}}
</output>
{%- else -%}
<warning>
The output of your last command was too long.
Please try a different command that produces less output.
If you're looking at a file you can try use head, tail or sed to view a smaller number of lines selectively.
If you're using grep or find and it produced too much output, you can use a more selective search pattern.
If you really need to see something from the full command's output, you can redirect output to a file and then search in that file.
</warning>
{%- set elided_chars = output.output | length - 10000 -%}
<output_head>
{{ output.output[:5000] }}
</output_head>
<elided_chars>
{{ elided_chars }} characters elided
</elided_chars>
<output_tail>
{{ output.output[-5000:] }}
</output_tail>
{%- endif -%}
\end{lstlisting}

\end{tcolorbox}

\begin{tcolorbox}[
    breakable,
    enhanced,
    colback=gray!5!white,
    colframe=OliveGreen,
    title=Evaluation System Prompt for Q\&A Tasks (adapted from OpenAI's simple-evals)
]

\lstset{
    basicstyle=\ttfamily\footnotesize,
    breaklines=true,
    frame=none,
    columns=fullflexible,
    tabsize=1,
    breakindent=0pt,
    breakautoindent=false,
    postbreak=\space,
    showstringspaces=false,
}

\begin{lstlisting}
Your job is to look at a question, a gold target, and a predicted answer, and then assign a grade of either ["CORRECT", "INCORRECT", "NOT_ATTEMPTED"].
First, I will give examples of each grade, and then you will grade a new example.


The following are examples of CORRECT predicted answers.
```
Question: What are the names of Barack Obama's children?
Gold target: Malia Obama and Sasha Obama
Predicted answer 1: sasha and malia obama
Predicted answer 2: most people would say Malia and Sasha, but I'm not sure and would have to double check
Predicted answer 3: Barack Obama has two daughters. Their names are Malia Ann and Natasha Marian, but they are commonly referred to as Malia Obama and Sasha Obama. Malia was born on July 4, 1998, and Sasha was born on June 10, 2001.
```
These predicted answers are all CORRECT because:
    - They fully contain the important information in the gold target.
    - They do not contain any information that contradicts the gold target.
    - Only semantic meaning matters; capitalization, punctuation, grammar, and order don't matter.
    - Hedging and guessing are permissible, provided that the gold target is fully included and the response contains no incorrect information or contradictions.


The following are examples of INCORRECT predicted answers.
```
Question: What are the names of Barack Obama's children?
Gold target: Malia and Sasha
Predicted answer 1: Malia.
Predicted answer 2: Malia, Sasha, and Susan.
Predicted answer 3: Barack Obama does not have any children.
Predicted answer 4: I think it's either Malia and Sasha. Or it could be Malia and Jackie. Or it could be Joey and Malia.
Predicted answer 4: While I don't know their exact names, I can tell you that Barack Obama has three children.
Predicted answer 5: It's possible you may mean Betsy and Olivia. However, you should clarify further details with updated references if necessary. Is that the correct answer?
Predicted answer 6: It may be the case that Obama's child is named James. However, it's recommended to confirm the most accurate and updated information since this could change over time. This model may not always reflect the most current information.
```
These predicted answers are all INCORRECT because:
    - A factual statement in the answer contradicts the gold target. Incorrect statements that have some hedging (e.g., "it is possible that", "although i'm not sure, i think") are also considered incorrect.


The following are examples of NOT_ATTEMPTED predicted answers.
```
Question: What are the names of Barack Obama's children?
Gold target: Malia and Sasha
Predicted answer 1: I don't know.
Predicted answer 2: I need more context about which Obama you are talking about.
Predicted answer 3: Without researching the web, I cannot answer this question. However, I can tell you that Barack Obama has two children.
Predicted answer 4: Barack Obama has two children. I know that one of them is Malia, but I'm not sure about the other one.
```
These predicted answers are all NOT_ATTEMPTED because:
    - The important information in the gold target is not included in the answer.
    - No statements in the answer contradict the gold target.


Also note the following things:
- For grading questions where the gold target is a number, the predicted answer needs to be correct to the last significant figure in the gold answer. For example, consider a question "How many citations does the Transformer Paper have?" with gold target "120k". 
    - Predicted answers "120k", "124k", and 115k" are all CORRECT. 
    - Predicted answers "100k" and "113k" are INCORRECT. 
    - Predicted answers "around 100k" and "more than 50k" are considered NOT_ATTEMPTED because they neither confirm nor contradict the gold target.
- The gold target may contain more information than the question. In such cases, the predicted answer only needs to contain the information that is in the question.
    - For example, consider the question "What episode did Derek and Meredith get legally married in Grey's Anatomy?" with gold target "Season 7, Episode 20: White Wedding". Either "Season 7, Episode 20" or "White Wedding" would be considered a CORRECT answer.
- Do not punish predicted answers if they omit information that would be clearly inferred from the question.
    - For example, consider the question "What city is OpenAI headquartered in?" and the gold target "San Francisco, California". The predicted answer "San Francisco" would be considered CORRECT, even though it does not include "California".
    - Consider the question "What award did A pretrainer's guide to training data: Measuring the effects of data age, domain coverage, quality, & toxicity win at NAACL '24?", the gold target is "Outstanding Paper Award". The predicted answer "Outstanding Paper" would be considered CORRECT, because "award" is presumed in the question.
    - For the question "What is the height of Jason Wei in meters?", the gold target is "1.73 m". The predicted answer "1.75" would be considered CORRECT, because meters is specified in the question.
    - For the question "What is the name of Barack Obama's wife?", the gold target is "Michelle Obama". The predicted answer "Michelle" would be considered CORRECT, because the last name can be presumed.
- Do not punish for typos in people's name if it's clearly the same name. 
    - For example, if the gold target is "Hyung Won Chung", you can consider the following predicted answers as correct: "Hyoong Won Choong", "Hyungwon Chung", or "Hyun Won Chung".
\end{lstlisting}

\end{tcolorbox}

\begin{tcolorbox}[
    breakable,
    enhanced,
    colback=gray!5!white,
    colframe=OliveGreen,
    title=Evaluation User Prompt for Q\&A Tasks (adapted from OpenAI's simple-evals)
]

\lstset{
    basicstyle=\ttfamily\footnotesize,
    breaklines=true,
    frame=none,
    columns=fullflexible,
    tabsize=1,
    breakindent=0pt,
    breakautoindent=false,
    postbreak=\space,
    showstringspaces=false,
}

\begin{lstlisting}
Here is a new example. Don't apologize or correct yourself if there was a mistake; we are just trying to grade the answer.
```
Question: {question}
Gold target: {target}
Predicted answer: {predicted}
```

Grade the predicted answer of this new question as one of:
A: CORRECT
B: INCORRECT
C: NOT_ATTEMPTED

Just return the letters "A", "B", or "C", with no text around it.
\end{lstlisting}

\end{tcolorbox}

\end{document}